\batchmode
\makeatletter
\def\input@path{{F:/Backups/Dropbox_20150901/Dropbox/Dropbox/self_Folder/myWorksOnDropboxs/201707_HyperUnmixingDatasets_ClassicAlgs//}}
\makeatother
\documentclass[twoside]{IEEEtran}
\usepackage[T1]{fontenc}
\usepackage{color}
\definecolor{page_backgroundcolor}{rgb}{1, 1, 1}
\pagecolor{page_backgroundcolor}
\usepackage{array}
\usepackage{rotating}
\usepackage{float}
\usepackage{url}
\usepackage{bm}
\usepackage{multirow}
\usepackage{amsmath}
\usepackage{amssymb}
\usepackage{graphicx}
\PassOptionsToPackage{normalem}{ulem}
\usepackage{ulem}
\usepackage[unicode=true,
 bookmarks=true,bookmarksnumbered=true,bookmarksopen=true,bookmarksopenlevel=1,
 breaklinks=true,pdfborder={0 0 0},backref=false,colorlinks=true]
 {hyperref}
\hypersetup{pdftitle={Your Title},
 pdfauthor={Your Name}}

\makeatletter

\providecommand{\tabularnewline}{\\}
\floatstyle{ruled}
\newfloat{algorithm}{tbp}{loa}
\providecommand{\algorithmname}{Algorithm}
\floatname{algorithm}{\protect\algorithmname}

\usepackage[caption=false,font=footnotesize]{subfig}
\usepackage{bm}
\usepackage{multirow}
\usepackage{algorithm}
\usepackage{algorithmic}
\usepackage{booktabs}% For formal tables
\usepackage{amssymb}
\usepackage[square,sort&compress,numbers]{natbib}

\@ifundefined{showcaptionsetup}{}{%
 \PassOptionsToPackage{caption=false}{subfig}}
\usepackage{subfig}
\makeatother

\begin{document}
\global\long\def\mtbfA{\mathbf{A}}
 \global\long\def\mtbfa{\mathbf{a}}
 \global\long\def\mebfA{\bar{\mtbfA}}
 \global\long\def\mebfa{\bar{\mtbfa}}

\global\long\def\mhbfA{\widehat{\mathbf{A}}}
 \global\long\def\mhbfa{\widehat{\mathbf{a}}}
 \global\long\def\mtcalA{\mathcal{A}}
 \global\long\def\mtbbA{\mathbb{A}}

\global\long\def\mtbfB{\mathbf{B}}
 \global\long\def\mtbfb{\mathbf{b}}
 \global\long\def\mebfB{\bar{\mtbfB}}
 \global\long\def\mebfb{\bar{\mtbfb}}

\global\long\def\mhbfB{\widehat{\mathbf{B}}}
 \global\long\def\mhbfb{\widehat{\mathbf{b}}}
 \global\long\def\mtcalB{\mathcal{B}}
 \global\long\def\mtbbB{\mathbb{B}}

\global\long\def\mtbfC{\mathbf{C}}
 \global\long\def\mtbfc{\mathbf{c}}
 \global\long\def\mebfC{\bar{\mtbfC}}
 \global\long\def\mebfc{\bar{\mtbfc}}

\global\long\def\mhbfC{\widehat{\mathbf{C}}}
 \global\long\def\mhbfc{\widehat{\mathbf{c}}}
 \global\long\def\mtcalC{\mathcal{C}}
 \global\long\def\mtbbC{\mathbb{C}}

\global\long\def\mtbfD{\mathbf{D}}
 \global\long\def\mtbfd{\mathbf{d}}
 \global\long\def\mebfD{\bar{\mtbfD}}
 \global\long\def\mebfd{\bar{\mtbfd}}

\global\long\def\mhbfD{\widehat{\mathbf{D}}}
 \global\long\def\mhbfd{\widehat{\mathbf{d}}}
 \global\long\def\mtcalD{\mathcal{D}}
 \global\long\def\mtbbD{\mathbb{D}}

\global\long\def\mtbfE{\mathbf{E}}
 \global\long\def\mtbfe{\mathbf{e}}
 \global\long\def\mebfE{\bar{\mtbfE}}
 \global\long\def\mebfe{\bar{\mtbfe}}

\global\long\def\mhbfE{\widehat{\mathbf{E}}}
 \global\long\def\mhbfe{\widehat{\mathbf{e}}}
 \global\long\def\mtcalE{\mathcal{E}}
 \global\long\def\mtbbE{\mathbb{E}}

\global\long\def\mtbfF{\mathbf{F}}
 \global\long\def\mtbff{\mathbf{f}}
 \global\long\def\mebfF{\bar{\mathbf{F}}}
 \global\long\def\mebff{\bar{\mathbf{f}}}

\global\long\def\mhbfF{\widehat{\mathbf{F}}}
 \global\long\def\mhbff{\widehat{\mathbf{f}}}
 \global\long\def\mtcalF{\mathcal{F}}
 \global\long\def\mtbbF{\mathbb{F}}

\global\long\def\mtbfG{\mathbf{G}}
 \global\long\def\mtbfg{\mathbf{g}}
 \global\long\def\mebfG{\bar{\mathbf{G}}}
 \global\long\def\mebfg{\bar{\mathbf{g}}}

\global\long\def\mhbfG{\widehat{\mathbf{G}}}
 \global\long\def\mhbfg{\widehat{\mathbf{g}}}
 \global\long\def\mtcalG{\mathcal{G}}
 \global\long\def\mtbbG{\mathbb{G}}

\global\long\def\mtbfH{\mathbf{H}}
 \global\long\def\mtbfh{\mathbf{h}}
 \global\long\def\mebfH{\bar{\mathbf{H}}}
 \global\long\def\mebfh{\bar{\mathbf{h}}}

\global\long\def\mhbfH{\widehat{\mathbf{H}}}
 \global\long\def\mhbfh{\widehat{\mathbf{h}}}
 \global\long\def\mtcalH{\mathcal{H}}
 \global\long\def\mtbbH{\mathbb{H}}

\global\long\def\mtbfI{\mathbf{I}}
 \global\long\def\mtbfi{\mathbf{i}}
 \global\long\def\mebfI{\bar{\mathbf{I}}}
 \global\long\def\mebfi{\bar{\mathbf{i}}}

\global\long\def\mhbfI{\widehat{\mathbf{I}}}
 \global\long\def\mhbfi{\widehat{\mathbf{i}}}
 \global\long\def\mtcalI{\mathcal{I}}
 \global\long\def\mtbbI{\mathbb{I}}

\global\long\def\mtbfJ{\mathbf{J}}
 \global\long\def\mtbfj{\mathbf{j}}
 \global\long\def\mebfJ{\bar{\mathbf{J}}}
 \global\long\def\mebfj{\bar{\mathbf{j}}}

\global\long\def\mhbfJ{\widehat{\mathbf{J}}}
 \global\long\def\mhbfj{\widehat{\mathbf{j}}}
 \global\long\def\mtcalJ{\mathcal{J}}
 \global\long\def\mtbbJ{\mathbb{J}}

\global\long\def\mtbfK{\mathbf{K}}
 \global\long\def\mtbfk{\mathbf{k}}
 \global\long\def\mebfK{\bar{\mathbf{K}}}
 \global\long\def\mebfk{\bar{\mathbf{k}}}

\global\long\def\mhbfK{\widehat{\mathbf{K}}}
 \global\long\def\mhbfk{\widehat{\mathbf{k}}}
 \global\long\def\mtcalK{\mathcal{K}}
 \global\long\def\mtbbK{\mathbb{K}}

\global\long\def\mtbfL{\mathbf{L}}
 \global\long\def\mtbfl{\mathbf{l}}
 \global\long\def\mebfL{\bar{\mathbf{L}}}
 \global\long\def\mebfl{\bar{\mathbf{l}}}

\global\long\def\mhbfL{\widehat{\mathbf{K}}}
 \global\long\def\mhbfl{\widehat{\mathbf{k}}}
 \global\long\def\mtcalL{\mathcal{L}}
 \global\long\def\mtbbL{\mathbb{L}}

\global\long\def\mtbfM{\mathbf{M}}
 \global\long\def\mtbfm{\mathbf{m}}
 \global\long\def\mebfM{\bar{\mathbf{M}}}
 \global\long\def\mebfm{\bar{\mathbf{m}}}

\global\long\def\mhbfM{\widehat{\mathbf{M}}}
 \global\long\def\mhbfm{\widehat{\mathbf{m}}}
 \global\long\def\mtcalM{\mathcal{M}}
 \global\long\def\mtbbM{\mathbb{M}}

\global\long\def\mtbfN{\mathbf{N}}
 \global\long\def\mtbfn{\mathbf{n}}
 \global\long\def\mebfN{\bar{\mathbf{N}}}
 \global\long\def\mebfn{\bar{\mathbf{n}}}

\global\long\def\mhbfN{\widehat{\mathbf{N}}}
 \global\long\def\mhbfn{\widehat{\mathbf{n}}}
 \global\long\def\mtcalN{\mathcal{N}}
 \global\long\def\mtbbN{\mathbb{N}}

\global\long\def\mtbfO{\mathbf{O}}
 \global\long\def\mtbfo{\mathbf{o}}
 \global\long\def\mebfO{\bar{\mathbf{O}}}
 \global\long\def\mebfo{\bar{\mathbf{o}}}

\global\long\def\mhbfO{\widehat{\mathbf{O}}}
 \global\long\def\mhbfo{\widehat{\mathbf{o}}}
 \global\long\def\mtcalO{\mathcal{O}}
 \global\long\def\mtbbO{\mathbb{O}}

\global\long\def\mtbfP{\mathbf{P}}
 \global\long\def\mtbfp{\mathbf{p}}
 \global\long\def\mebfP{\bar{\mathbf{P}}}
 \global\long\def\mebfp{\bar{\mathbf{p}}}

\global\long\def\mhbfP{\widehat{\mathbf{P}}}
 \global\long\def\mhbfp{\widehat{\mathbf{p}}}
 \global\long\def\mtcalP{\mathcal{P}}
 \global\long\def\mtbbP{\mathbb{P}}

\global\long\def\mtbfQ{\mathbf{Q}}
 \global\long\def\mtbfq{\mathbf{q}}
 \global\long\def\mebfQ{\bar{\mathbf{Q}}}
 \global\long\def\mebfq{\bar{\mathbf{q}}}

\global\long\def\mhbfQ{\widehat{\mathbf{Q}}}
 \global\long\def\mhbfq{\widehat{\mathbf{q}}}
\global\long\def\mtcalQ{\mathcal{Q}}
 \global\long\def\mtbbQ{\mathbb{Q}}

\global\long\def\mtbfR{\mathbf{R}}
 \global\long\def\mtbfr{\mathbf{r}}
 \global\long\def\mebfR{\bar{\mathbf{R}}}
 \global\long\def\mebfr{\bar{\mathbf{r}}}

\global\long\def\mhbfR{\widehat{\mathbf{R}}}
 \global\long\def\mhbfr{\widehat{\mathbf{r}}}
\global\long\def\mtcalR{\mathcal{R}}
 \global\long\def\mtbbR{\mathbb{R}}

\global\long\def\mtbfS{\mathbf{S}}
 \global\long\def\mtbfs{\mathbf{s}}
 \global\long\def\mebfS{\bar{\mathbf{S}}}
 \global\long\def\mebfs{\bar{\mathbf{s}}}

\global\long\def\mhbfS{\widehat{\mathbf{S}}}
 \global\long\def\mhbfs{\widehat{\mathbf{s}}}
\global\long\def\mtcalS{\mathcal{S}}
 \global\long\def\mtbbS{\mathbb{S}}

\global\long\def\mtbfT{\mathbf{T}}
 \global\long\def\mtbft{\mathbf{t}}
 \global\long\def\mebfT{\bar{\mathbf{T}}}
 \global\long\def\mebft{\bar{\mathbf{t}}}

\global\long\def\mhbfT{\widehat{\mathbf{T}}}
 \global\long\def\mhbft{\widehat{\mathbf{t}}}
 \global\long\def\mtcalT{\mathcal{T}}
 \global\long\def\mtbbT{\mathbb{T}}

\global\long\def\mtbfU{\mathbf{U}}
 \global\long\def\mtbfu{\mathbf{u}}
 \global\long\def\mebfU{\bar{\mathbf{U}}}
 \global\long\def\mebfu{\bar{\mathbf{u}}}

\global\long\def\mhbfU{\widehat{\mathbf{U}}}
 \global\long\def\mhbfu{\widehat{\mathbf{u}}}
 \global\long\def\mtcalU{\mathcal{U}}
 \global\long\def\mtbbU{\mathbb{U}}

\global\long\def\mtbfV{\mathbf{V}}
 \global\long\def\mtbfv{\mathbf{v}}
 \global\long\def\mebfV{\bar{\mathbf{V}}}
 \global\long\def\mebfv{\bar{\mathbf{v}}}

\global\long\def\mhbfV{\widehat{\mathbf{V}}}
 \global\long\def\mhbfv{\widehat{\mathbf{v}}}
\global\long\def\mtcalV{\mathcal{V}}
 \global\long\def\mtbbV{\mathbb{V}}

\global\long\def\mtbfW{\mathbf{W}}
 \global\long\def\mtbfw{\mathbf{w}}
 \global\long\def\mebfW{\bar{\mathbf{W}}}
 \global\long\def\mebfw{\bar{\mathbf{w}}}

\global\long\def\mhbfW{\widehat{\mathbf{W}}}
 \global\long\def\mhbfw{\widehat{\mathbf{w}}}
 \global\long\def\mtcalW{\mathcal{W}}
 \global\long\def\mtbbW{\mathbb{W}}

\global\long\def\mtbfX{\mathbf{X}}
 \global\long\def\mtbfx{\mathbf{x}}
 \global\long\def\mebfX{\bar{\mtbfX}}
 \global\long\def\mebfx{\bar{\mtbfx}}

\global\long\def\mhbfX{\widehat{\mathbf{X}}}
 \global\long\def\mhbfx{\widehat{\mathbf{x}}}
 \global\long\def\mtcalX{\mathcal{X}}
 \global\long\def\mtbbX{\mathbb{X}}

\global\long\def\mtbfY{\mathbf{Y}}
 \global\long\def\mtbfy{\mathbf{y}}
\global\long\def\mebfY{\bar{\mathbf{Y}}}
 \global\long\def\mebfy{\bar{\mathbf{y}}}

\global\long\def\mhbfY{\widehat{\mathbf{Y}}}
 \global\long\def\mhbfy{\widehat{\mathbf{y}}}
 \global\long\def\mtcalY{\mathcal{Y}}
 \global\long\def\mtbbY{\mathbb{Y}}

\global\long\def\mtbfZ{\mathbf{Z}}
 \global\long\def\mtbfz{\mathbf{z}}
 \global\long\def\mebfZ{\bar{\mathbf{Z}}}
 \global\long\def\mebfz{\bar{\mathbf{z}}}

\global\long\def\mhbfZ{\widehat{\mathbf{Z}}}
 \global\long\def\mhbfz{\widehat{\mathbf{z}}}
\global\long\def\mtcalZ{\mathcal{Z}}
 \global\long\def\mtbbZ{\mathbb{Z}}

\global\long\def\mtth{\text{th}}

\global\long\def\mtbfzero{\mathbf{0}}
 \global\long\def\mtbfone{\mathbf{1}}

\global\long\def\mttrace{\text{Tr}}

\global\long\def\mttotalVariation{\text{TV}}

\global\long\def\mtexpect{\mathbb{E}}

\global\long\def\mtdet{\text{det}}

\global\long\def\mtvec{\mathbf{\text{vec}}}

\global\long\def\mtvar{\mathbf{\text{var}}}

\global\long\def\mtcov{\mathbf{\text{cov}}}

\global\long\def\mtsubTo{\mathbf{\text{s.t.}}}

\global\long\def\mtfor{\text{for}}

\global\long\def\mtrank{\text{rank}}

\global\long\def\mtrankn{\text{rankn}}

\global\long\def\mtdiag{\mathbf{\text{diag}}}

\global\long\def\mtsign{\mathbf{\text{sign}}}

\global\long\def\mtloss{\mathbf{\text{loss}}}

\global\long\def\mtwhen{\text{when}}

\global\long\def\mtwhere{\text{where}}

\global\long\def\mtif{\text{if}}

\global\long\def\mtbfM{\mathbf{M}}
 \global\long\def\mtbfm{\mathbf{m}}
 \global\long\def\mtbfY{\mathbf{Y}}
 \global\long\def\mtbfy{\mathbf{y}}

\global\long\def\mtbfA{\mathbf{A}}
 \global\long\def\mtbfa{\mathbf{a}}
 \global\long\def\mtbfE{\mathbf{E}}
 \global\long\def\mtbfe{\mathbf{e}}

\global\long\def\mtbfC{\mathbf{C}}
 \global\long\def\mtbfc{\mathbf{c}}
\global\long\def\mtbfU{\mathbf{U}}
 \global\long\def\mtbfu{\mathbf{u}}

\global\long\def\mtbfW{\mathbf{W}}
 \global\long\def\mtbfD{\mathbf{D}}
 \global\long\def\mtbfL{\mathbf{L}}
 \global\long\def\mtbfz{\mathbf{z}}

\global\long\def\mtbldM{\mathbf{M}}
 \global\long\def\mtbldm{\mathbf{m}}
 \global\long\def\mtbldY{\mathbf{Y}}
 \global\long\def\mtbldy{\mathbf{y}}

\global\long\def\mtbldA{\mathbf{A}}
 \global\long\def\mtblda{\mathbf{a}}
 \global\long\def\mtbldE{\mathbf{E}}
 \global\long\def\mtblde{\mathbf{e}}

\global\long\def\mtbldW{\mathbf{W}}
 \global\long\def\mtbldD{\mathbf{D}}
 \global\long\def\mtbldL{\mathbf{L}}
 \global\long\def\mtbldz{\mathbf{z}}

\global\long\def\tensorY{\mathcal{Y}}
 \global\long\def\tensorA{\mathcal{A}}
 \global\long\def\inSetsR{\in\mathbb{R}}

\title{Hyperspectral Unmixing: Ground Truth Labeling, Datasets, Benchmark
Performances and Survey}

\author{Feiyun Zhu}
\maketitle
\begin{abstract}
Hyperspectral unmixing (HU) is a very useful and increasingly popular
preprocessing step for a wide range of hyperspectral applications.
However, the HU research has been constrained a lot by three factors:
(a) the number of hyperspectral images (especially the ones with ground
truths) are very limited; (b) the ground truths of most hyperspectral
images are not shared on the web, which may cause lots of unnecessary
troubles for researchers to evaluate their algorithms; (c) the codes
of most state-of-the-art methods are not shared, which may also delay
the testing of new methods. 

Accordingly, this paper deals with the above issues from the following
three perspectives: (1) as a profound contribution, we provide a general
labeling method for the HU. With it, we labeled up to 15 hyperspectral
images, providing 18 versions of ground truths. To the best of our
knowledge, this is the first paper to summarize and share up to 15
hyperspectral images and their 18 versions of ground truths for the
HU. Observing that the hyperspectral classification (HyC) has much
more standard datasets (whose ground truths are generally publicly
shared) than the HU, we propose an interesting method to transform
the HyC datasets for the HU research. (2) To further facilitate the
evaluation of HU methods under different conditions, we reviewed and
implemented the algorithm to generate a complex synthetic hyperspectral
image. By tuning the hyper-parameters in the code, we may verify the
HU methods from four perspectives. The code would also be shared on
the web. (3) To provide a standard comparison, we reviewed up to 10
state-of-the-art HU algorithms, then selected the 5 most benchmark
HU algorithms, and compared them on the 15 real hyperspectral datasets.
The experiment results are surely reproducible; the implemented codes
would be shared on the web.\end{abstract}


\section{Introduction}

\begin{figure*}[t]
\noindent \begin{centering}
\includegraphics[width=1.5\columnwidth]{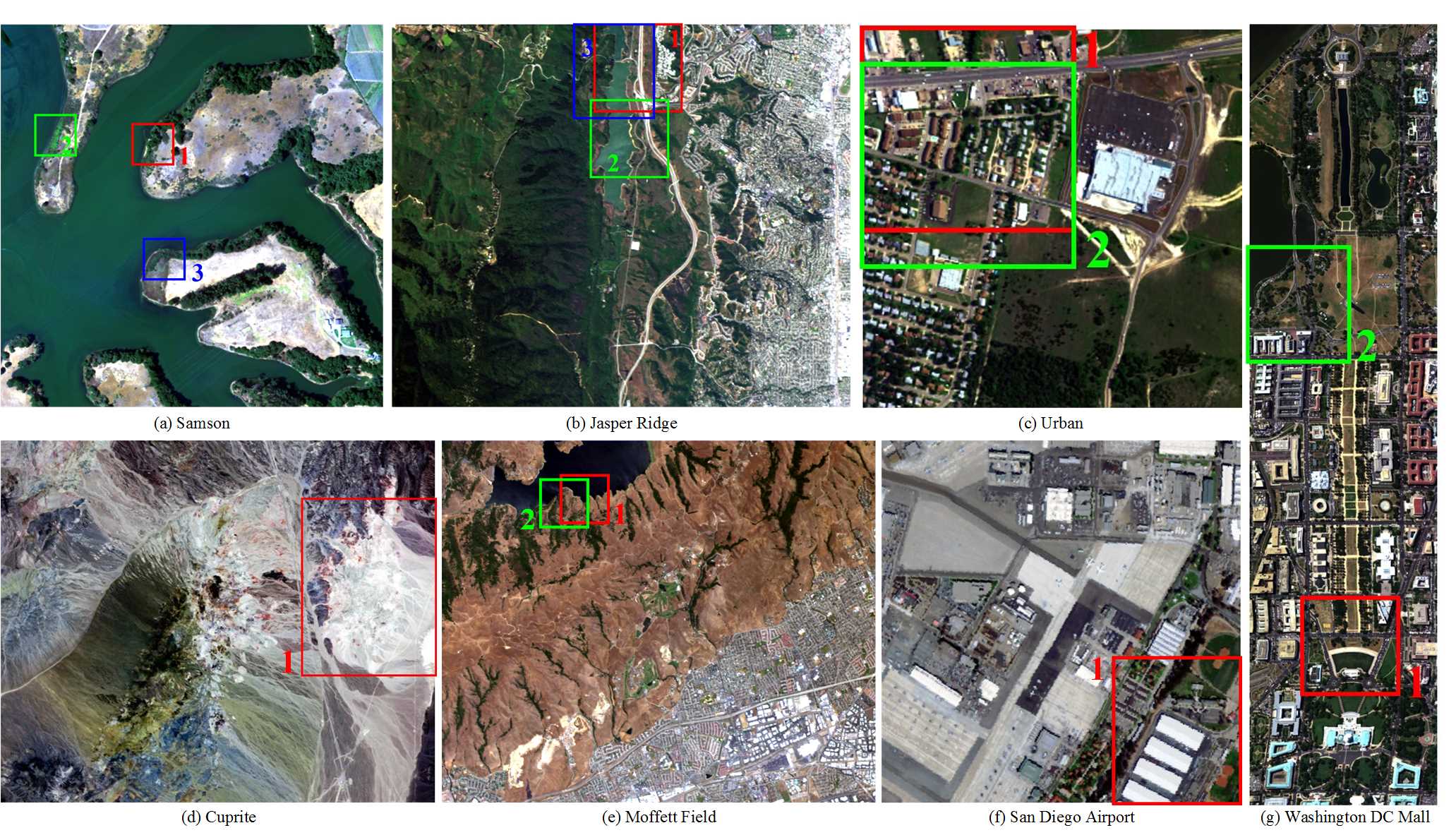}
\par\end{centering}

\caption{The illustration of 7 popular hyperspectral images where we select
15 subimages (i.e., ROIs) and provide 18 versions of ground truths
for the HU study. They are \textbf{{[}a{]}} Samson (3 ROIs), \textbf{{[}b{]}}
Jasper Ridge (3 ROIs), \textbf{{[}c{]}} Urban (the full image and
2 ROIs), \textbf{{[}d{]}} Cuprite (1 ROI), \textbf{{[}e{]}} Moffett
Field (2 ROIs), \textbf{{[}f{]}} San Diego Airport (1 ROI), and \textbf{{[}g{]}}
Washington DC Mall (2 ROIs). In Section\ \ref{sec:Real-Hyperspectral-Images}
and Table\ \ref{tab:18_realHyperDatasets}, we provide very detailed
information about the 15 ROIs. \label{fig:7_hyperImages}}
\end{figure*}
\IEEEPARstart{H}{yperspectral} remote sensing has been widely used
in lots of applications\footnote{such as mining \& oil industries, agriculture, food safety, pharmaceutical
process monitoring and quality control, biomedical \& biometric, surveillance,
military, environment monitoring and forensic applications etc.} since it can capture a 3D image cube at hundreds of contiguous bands
across the electromagnetic spectrum, providing substantial information
of the scene\ \cite{Dias_12_AEORS_HUoverview}. However, due to the
microscopic material mixing, multiple scattering and low spatial resolution
of hyperspectral sensors, the pixel spectra are inevitably mixed with
various substances\ \cite{yingWang_2015_TIP_RobustUnmixing,Dias_12_AEORS_HUoverview},
resulting in lots of mixed pixels. Accordingly, hyperspectral unmixing
(HU) is an essential processing step for various hyperspectral image
applications, such as high-resolution hyperspectral imaging~\cite{Kawakami_2011_CVPR_highResolutionHyperImaging,Lanaras_2015_ICCV_superResolution,dong2016hyperspectral,ma2014acquisition,akhtar2014sparse,akhtar2015bayesian},
hyperspectral enhancement~\cite{zhGuo_2009_SPIE_L1unmixing&Enhancement},
sub-pixel mapping~\cite{Mertens_2003_IJRS_subPixelMapping}, hyperspectral
compression and reconstruction~\cite{cbLi_2012_Tip_CompressSensing&Unmixing},
detection and identification substances in the scene~\cite{zhGuo_2009_SPIE_L1unmixing&Enhancement,Qian_11_TGRS_NMF+l1/2},
hyperspectral visualization~\cite{ShangshuCai_2007_TGRS_hyperVisualizationUsingDoubleLayers,fyzhu_2014_TIP_DgS_NMF},
etc. 

\begin{table*}[t]
\centering{}\caption{The 15 real hyperspectral (sub-) images and their information, including
the image size, the numbers of all spectral bands and the selected
bands, the \emph{endmembers} numbers, the illustration of scenes and
their 18 versions of ground truths etc. \label{tab:18_realHyperDatasets}}
\vspace{-0.25cm}
 \resizebox{1.95\columnwidth}{!}{%
\begin{tabular}{l|cc|cc|c|c|c|c}
\hline 
\multirow{2}{*}{Real Datasets} &
\multicolumn{2}{c|}{Image sizes} &
\multicolumn{2}{c|}{Number of bands} &
\# End- &
Ground truth &
The top-left pixel in &
Subimages\tabularnewline
\cline{2-5} 
 & \# row &
\# column &
\# all bands &
\# selected bands &
member &
shown in &
original image &
shown in \tabularnewline
\hline 
1. Samson\#1 &
95 &
95  &
\multirow{3}{*}{156} &
\multirow{3}{*}{156 } &
3 &
Fig.\,\ref{fig:GT_Samson=0000231} &
$\left(252,\,332\right)$  &
\multirow{3}{*}{Fig.\,\ref{fig:7_hyperImages}a}\tabularnewline
2. Samson\#2 &
95 &
95  &
 &  & 3 &
\multirow{1}{*}{Fig.\,\ref{fig:GT_Samson=0000232}} &
$\left(232,\,93\right)$ & \tabularnewline
3. Samson\#3 &
95 &
95  &
 &  & 3 &
Fig.\,\ref{fig:GT_Samson=0000233} &
$\left(634,\,455\right)$ & \tabularnewline
\hline 
4. Jasper Ridge\#1 &
115 &
115 &
\multirow{3}{*}{224} &
\multirow{3}{*}{198} &
5 &
Fig.\,\ref{fig:GT_JasperRidge=0000231} &
$\left(1,\,272\right)$  &
\multirow{3}{*}{Fig.\,\ref{fig:7_hyperImages}b}\tabularnewline
5. Jasper Ridge\#2 &
100 &
100 &
 &  & 4 &
Fig.\,\ref{fig:GT_JasperRidge=0000232} &
$\left(105,\,269\right)$ & \tabularnewline
6. Jasper Ridge\#3 &
122 &
104 &
 &  & 4 &
Fig.\,\ref{fig:GT_JasperRidge=0000233} &
$\left(1,\,246\right)$ & \tabularnewline
\hline 
7,\,8,\,9. Urban &
307 &
307 &
\multirow{3}{*}{221} &
\multirow{3}{*}{162} &
4,\,5,\,6 &
Figs.\,\ref{fig:GT_Urban}a,\,\ref{fig:GT_Urban}b,\,\ref{fig:GT_Urban}c &
$\left(1,\,1\right)$ &
\multirow{3}{*}{Fig.\,\ref{fig:7_hyperImages}c}\tabularnewline
10. Urban\#1 &
160 &
168 &
 &  & 6 &
Fig.\,\ref{fig:GT_Urban=0000231} &
$\left(1,\,1\right)$ & \tabularnewline
11. Urban\#2 &
160 &
168 &
 &  & 6 &
Fig.\,\ref{fig:GT_Urban=0000232} &
$\left(40,\,1\right)$ & \tabularnewline
\hline 
12. Cuprite &
250 &
190 &
224 &
188 &
12 &
Fig.\,\ref{fig:GT_Cuprite} &
$\left(86,\,425\right)$ &
Fig.\,\ref{fig:7_hyperImages}d\tabularnewline
\hline 
13. Moffett Field\#1 &
60 &
60 &
\multirow{2}{*}{224} &
\multirow{2}{*}{196} &
3 &
Fig.\,\ref{fig:GT_Moffett Field=0000231} &
$\left(54,\,172\right)$ &
\multirow{2}{*}{Fig.\,\ref{fig:7_hyperImages}e}\tabularnewline
14. Moffett Field\#2 &
60 &
60 &
 &  & 3 &
Fig.\,\ref{fig:GT_Moffett Field=0000232} &
$\left(62,\,146\right)$ & \tabularnewline
\hline 
15,\,16. San Deigo Airport &
160 &
140 &
224 &
189 &
4,\,5 &
Figs.\,\ref{fig:GT_SanDiegoAirport}a,\,\ref{fig:GT_SanDiegoAirport}b &
$\left(241,\,261\right)$ &
Fig.\,\ref{fig:7_hyperImages}f\tabularnewline
\hline 
17. Washington DC Mall\#1 &
150 &
150 &
\multirow{2}{*}{224?} &
\multirow{2}{*}{191} &
6 &
Fig.\,\ref{fig:Washington-DC-Mall=0000231} &
$\left(549,\,160\right)$ &
\multirow{2}{*}{Fig.\,\ref{fig:7_hyperImages}g}\tabularnewline
18. Washington DC Mall\#2 &
180 &
160 &
 &  & 6 &
Fig.\,\ref{fig:Washington-DC-Mall=0000232} &
$\left(945,\,90\right)$ & \tabularnewline
\hline 
\end{tabular} }
\end{table*}
 The HU aims to decompose each (mixed) pixel into a set of ``pure''
pixels (called \emph{endmembers} such as the spectra of grass, water
etc.), weighted by the corresponding proportions, called \emph{abundances}\ \cite{fyzhu_2014_IJPRS_SSNMF,fyzhu_2014_TIP_DgS_NMF,fyzhu_2014_JSTSP_RRLbS}.
Formally, given a hyperspectral image with $L$ channels and $N$
pixels, each pixel $\mtbfy\in\mtbbR_{+}^{L}$ is assumed to be a composite
of $K$ \emph{endmembers} $\left\{ \mtbfm_{k}\right\} _{k=1}^{K}\in\mtbbR_{+}^{L}$.
The linear combinatorial model is the most commonly used one 
\begin{equation}
\mtbfx=\sum_{k=1}^{K}\mtbfm_{k}a_{k},\quad\mtsubTo\:a_{k}\geq0\ \text{and\ }\sum_{k=1}^{K}a_{k}=1,\label{eq:LinearMixedModel_vector}
\end{equation}
where $a_{k}$ is the composite \emph{abundance} of the $k^{\mtth}$
\emph{endmember}. In the unsupervised setting, both \emph{endmembers}
$\left\{ \mtbfm_{k}\right\} _{k=1}^{K}$ and \emph{abundances} $\left\{ a_{k}\right\} _{k=1}^{K}$
are unknown, which makes the HU a challenging problem\ \cite{guangliangCheng_2014_ICIP,guangliangCheng_2015_ICIP,guangliangCheng_2016_neurocomputing,guangliangCheng_2016_JStars_robustHyperClassification,guangliangCheng_2016_TGRSL,yaoyao_2017_MICCAI,xinliang_2017_CVPR_WSISA,zhengxu_2017_ACMBCB,fyzhu_2017_RLDM_WarmStart,fyZhu_2017_arXiv_CohesionDrivenActorCriticRL}.

The HU is a very hot research topic\textemdash tens or even hundreds
HU papers have been published each year. However, the HU research
has been constrained a lot since the commonly used datasets (especially
their ground truths), are generally not shared on the web. Such case
will surely hinder the development of new methods; the researchers,
who are interested in the HU, will have to make great efforts to do
the preparation works\textemdash they have to find the hyperspectral
datasets and their ground truths, which generally ends up with failures.
Instead, they have to label the hyperspectral images, which is very
challenging and needs lots of techniques. 

In order to promote the HU research, we write this paper from the
following five perspectrives:

\textbf{1)}. \uline{This is the first paper to introduce a general
method to label hyperspectral images for the HU research}. We provide
the method to label the \emph{endmembers} \& \emph{abundances} (in
Sections\ \ref{sub:The-Endmember-Labelling},\,\ref{sub:The-Abundance-Labelling}),
as well as the method to evaluate the labeling results (in Section\ \ref{sub:verify_labelling_result}).
The hyperspectral classification (HyC) is similar to the HU task;
it has much more standard datasets with publicly available ground
truths. We propose an interesting method to transform the benchmark
HyC datasets for the HU research. Please refer to Section\ \ref{sec:the method to label GT 4 HU}.

\textbf{2)}. Moreover, we are the first to \uline{summarize the
information of 15 most commonly used hypserpectral images as well
as their 18 versions of ground truths} for the HU. All of them will
be shared on the web as a standard dataset for the evaluation of new
HU methods (cf. Section\ \ref{sec:Real-Hyperspectral-Images}, Table\,\ref{tab:18_realHyperDatasets}
and Fig.\ \ref{fig:7_hyperImages}).

\textbf{3)}. We reviewed and implemented the method to \uline{generate
a complex synthetic hyperspectral image} for the HU research in Section\ \ref{sec:simulated_hyperspectral_images}.
This synthetic image is widely used in the papers\ \cite{Miao_07_ITGRS_NMFMVC,Jia_09_TGRS_ConstainedNMF,yingWang_2015_TIP_RobustUnmixing,Qian_11_TGRS_NMF+l1/2,wang2016parallel,tong2016nonnegative,tong2017region,qian2017matrix}.
The code will be also shared on the web. 

\textbf{4)}. We \uline{reviewed 10 state-of-the-art methods}, summarized
their main ideas and algorithms as well as pointed out their internal
relations. Besides, we provide the codes of 5 successful state-of-the-art
HU methods. Please refer to Section\ \ref{sec:Review_NMF=000026Extensions}.

\textbf{5)}. The \uline{5 most popular methods} are implemented
and \uline{compared on the 15 real hyperspectral images}. We provide
their results as the benchmark HU performance in Section\,\ref{sec:ExpRsts}.

\section{The three categories of HU methods\label{sec:3_types_methods}}

In general, existing HU methods can be classified into three categories:
supervised methods\,\cite{zhGuo_2009_SPIE_L1unmixing&Enhancement,cbLi_2012_Tip_CompressSensing&Unmixing},
weakly supervised methods\,\cite{Iordache_2011_TGRS_SparseUnmixing,Iordache_2012_TGRS_TV_SparseUnmixing}
and unsupervised methods\,\cite{Bayliss_1997_SPIE_ICA_unmixing,SJia_07_TGRS_SSCBSS,Jia_09_TGRS_ConstainedNMF,Qian_11_TGRS_NMF+l1/2,nWang_13_SelectedTopics_EDC-NMF,fyzhu_2014_IJPRS_SSNMF}\footnote{Here supervise, weakly supervise and supervise are totally different
from the those terms in the general machine learning.}. The \emph{endmember} is given beforehand in the supervised setting;
only the \emph{abundance} is required to estimate. Although, the HU
task is simplified in this setting, it is usually intractable to obtain
feasible \emph{endmembers}, thus, hampering the acquisition of good
HU estimations. 

Accordingly, the weakly supervised methods\,\cite{Iordache_2011_TGRS_SparseUnmixing,Iordache_2012_TGRS_TV_SparseUnmixing}
were proposed. A large library of material spectra had been collected
by a field spectrometer beforehand\footnote{e.g., the United States Geological Survey (USGS) digital spectral
library.}. Then, the HU task becomes the problem of finding an optimal subset
of material spectra in the library that can best represent all the
pixels in the hyperspectral image\ \cite{Iordache_2011_TGRS_SparseUnmixing}
as well as their \emph{abundance} maps. Unfortunately, the library
is far from optimal because the spectra in it are not standardly unified.
First, for different hyperspectral sensors, the spectral signatures
of the same material can be very inconsistent. Second, for the hyperspectral
images recorded by different sensors, both the number of spectral
bands and the electromagnetic range of recorded spectra can be largely
different as well\textemdash for example, some images (like Samson)
have $156$ channels covering the spectra from $401\ nm$ to $889\ nm$,
while other images (like Cuprite and Jasper Ridge) have $224$ channels
covering the spectra from $370\ nm$ to $2,480\ nm$. Finally, the
recording conditions are very different\textemdash some hyperspectral
images are captured far from the outer space, while some hyperspectral
images are obtained from the airplane or even in the lab. Due to the
atmospheric effects etc., the different recording conditions would
result in different spectral appearances. In short, the weakness of
the spectral library brings side effects into this kind of methods. 

More commonly, the \emph{endmembers} are learned from the hyperspectral
image itself to ensure the spectral coherence\,\cite{zhGuo_2009_SPIE_L1unmixing&Enhancement}---the
unsupervised HU methods are preferred, where both the \emph{endmember}
and \emph{abundances} are learned from the hyperspectral image. Specifically,
unsupervised HUs can be  categorized into two types: geometric methods~\cite{Boardman_1995_JPL_PPI,Jose_05_TGRS_Vca,Chang_2006_TGRS_SGA,junLi_2008_IGARSS_MVSA,Bioucas_2009_WHISPERS_SISAL,Martin_12_SelectedTopics_SSPP}
and statistical ones~\cite{Wang_06_ITGRS_ICAextEndmember,Dobigeon_ISP_BaysianHU,Jose_09_WHISPERS_SplittingAugmentedLag,Jose_12_TGRS_HUbyDirchletComponents,Chanussot_2014_TGRS_NonlinearUnmixing}.
The geometric methods usually exploit the simplex to model the distribution
of pixel spectra.  The N-FINDR~\cite{Michael_99_PSCIS_nFindr}
and Vertex Component Analysis (VCA)~\cite{Jose_05_TGRS_Vca} are
the most benchmark geometric methods. For the N-FINDR, the \emph{endmembers}
are extracted by inflating a simplex inside the hyperspectral pixel
space and treating the vertices of a simplex with the largest volume
as \emph{endmembers}~\cite{Michael_99_PSCIS_nFindr}. The VCA~\cite{Jose_05_TGRS_Vca}
projects all the residual pixel onto a direction orthogonal to the
simplex spanned by the chosen\emph{ endmembers}; the new \emph{endmember}
is identified as the extreme of the projection. Although these methods
are simple and fast, they suffer from the requirement of pure pixels,
which is usually unreliable in practice\,\cite{Jia_09_TGRS_ConstainedNMF,Qian_11_TGRS_NMF+l1/2,XLu_2013_TGRS_ManifoldSparseNMF}. 

Accordingly, many statistical methods have been proposed for or applied
to the HU problem, among which the Nonnegative Matrix Factorization
(NMF)~\cite{Lee_99_Nature_NMF} and its extensions are the most popular.
In (the following) Section\ \ref{sec:Review_NMF=000026Extensions},
we will review the NMF and its extensions in detail.

\section{Review of the state-of-the-art HU methods\label{sec:Review_NMF=000026Extensions}}

\textbf{Notations}. For clear modeling, the boldface uppercase letter
(e.g. $\mtbfX$) and lowercase letter ($\mtbfx$) are used to represent
matrices and vectors respectively. Given a matrix $\mtbfX\triangleq\left\{ X_{ln}\right\} \in\mtbbR^{L\times N}$,
$\mtbfx^{l}\in\mtbbR^{1\times N}$ is the $l^{\mtth}$ row vector
and $\mtbfx_{n}\in\mtbbR^{L}$ denotes the $n^{\mtth}$ column vector.
$X_{ln}$ is the $\left(l,n\right)$-th element in the matrix. $\mtbfX\geq\mtbfzero$
or $\mtbfX\in\mtbbR_{+}^{L\times N}$ represent a nonnegative matrix.
The $\ell_{2,1}$-norm of matrices is defined as $\left\Vert \mtbfX\right\Vert _{2,1}=\sum_{l}^{L}\left(\sum_{n}^{N}X_{ln}^{2}\right)^{1/2}$. 

\textbf{Formalization}. In the HU modeling, the hyperspectral image
with $L$ channles (or bands) and $N$ pixels, is represented by a
nonnegative matrix $\mtbfX\triangleq\left[\mtbfx_{1},\mtbfx_{2},\cdots,\mtbfx_{N}\right]\in\mtbbR_{+}^{L\times N}$.
From the perspective of the linear mixture assumption, the goal of
HU is to find two nonnegative matrices to well approximate $\mtbfX$
with their product: 
\begin{align}
\min_{\mtbfM,\mtbfA}\  & \mtloss\left\{ \mtbfX,\widetilde{\mtbfX}\right\} +\lambda\Psi\left(\mtbfA\right)+\alpha\Phi\left(\mtbfM\right),\label{eq:generalized_HU_loss}\\
\mtsubTo & \ \widetilde{\mtbfX}=\mtbfM\mtbfA,\mtbfM\geq\mtbfzero,\mtbfA\geq\mtbfzero,\nonumber 
\end{align}
where $\widetilde{\mtbfX}$ is the approximation of the original hyperspectral
image; $\mtbfM\triangleq\left[\mtbfm_{1},\cdots,\mtbfm_{K}\right]\in\mtbbR_{+}^{L\times K}$
is the \emph{endmember} matrix that consists of $K$ pure pixel spectra
and $K\ll\min\left\{ L,N\right\} $; $\mtbfA\triangleq\left[\mtbfa_{1},\cdots,\mtbfa_{N}\right]\in\mtbbR_{+}^{K\times N}$
is the\emph{ abundance} matrix\textemdash the $n^{\mtth}$ column
vector $\mtbfa_{n}$ contains all the $K$ \emph{abundances} at pixel
$\mtbfx_{n}$; $\mtloss\left\{ \cdot,\cdot\right\} $ is a loss function
measuring the difference between two terms; $\Psi\left(\mtbfA\right)$
is some kind of constraints on the \emph{abundance} maps; $\Phi\left(\mtbfM\right)$
is the constraint on the \emph{endmembers}.

\subsection{Nonnegative Matrix Factorization (i.e., NMF)\ \cite{Lee_99_Nature_NMF,Lee_00_NIPS_NMF}\label{sub:Nonnegative-Matrix-Factorization}}

When the $\mtloss\left\{ \cdot,\cdot\right\} $ is the Euclidean loss,
the objective\ \eqref{eq:generalized_HU_loss} becomes the standard
NMF problem\ \cite{fyzhu_2014_TIP_DgS_NMF,Lee_99_Nature_NMF,Lee_00_NIPS_NMF},
which is commonly used in a wide range of applications including the
HU\ \cite{LiuXueSong_2011_TGRS_ConstrainedNMF,Miao_07_ITGRS_NMFMVC,Jia_09_TGRS_ConstainedNMF,fyzhu_2014_TIP_DgS_NMF,Qian_11_TGRS_NMF+l1/2,nWang_13_SelectedTopics_EDC-NMF}.
Since\ \eqref{eq:generalized_HU_loss} is non-convex w.r.t. the two
variables (i.e., $\mtbfM$ and $\mtbfA$) together\ \cite{Lee_99_Nature_NMF,Lee_00_NIPS_NMF},
it is unrealistic to find global minima. Alternatively, Lee and Seung\ \cite{Lee_99_Nature_NMF,Lee_00_NIPS_NMF}
have proposed the multiplicative update rules as follows:
\begin{equation}
M_{lk}\leftarrow\frac{M_{lk}\left(\mtbfX\mtbfA^{\top}\right)_{lk}}{\left(\mtbfM\mtbfA\mtbfA^{\top}\right)_{lk}},\quad A_{kn}\leftarrow\frac{A_{kn}\left(\mtbfM^{\top}\mtbfX\right)_{kn}}{\left(\mtbfM^{\top}\mtbfM\mtbfA\right)_{kn}},\label{eq:NMF_updating_M=000026A}
\end{equation}
 which have been proved to be non-increasing. Apart from\ \eqref{eq:NMF_updating_M=000026A},
there are other optimization algorithms to solve the problem\ \eqref{eq:generalized_HU_loss},
such as the active-set\ \cite{Kim_08_JMAA_activeSetNMF}, the alternation
nonnegative least least squares \cite{Berry_2007_CSDA_NMF_emailSurveillance}
and the projected gradient descent \cite{Lin_2007_neuralCompation_PGM_NMF}.

Although NMF is well adapted to face analyses~\cite{Stanzli_01_CVPR_locNMF,Roman_11_PAMI_EarthMoverNMF}
and documents clustering~\cite{WeiXu_03_SIGIR_DocClusterNMF,Farial_06_InfProManag_DocumNMF},
the objective function\ \eqref{eq:generalized_HU_loss} is non-convex,
naturally resulting in a large solution space~\cite{Lee_00_NIPS_NMF}.
Many extension methods have been proposed by employing suitable priors
to restrict the solution space. For the HU task, the priors are imposed
either on \emph{abundances}\ \cite{Qian_11_TGRS_NMF+l1/2,JmLiu_12_SlectedTopics_W-NMF,LiuXueSong_2011_TGRS_ConstrainedNMF}
(cf. Section\ \ref{sub:NMF_Constrained_Abundance}) or on\emph{ endmembers}\ \cite{Miao_07_ITGRS_NMFMVC,nWang_13_SelectedTopics_EDC-NMF}
(cf. Section\ \ref{sub:NMF_Constrained_endmembers}).

\subsection{The NMF extensions with constraints on the abundances\emph{\label{sub:NMF_Constrained_Abundance}\label{sub:NMF-extensions_abundances}}}

This section reviews the NMF extensions that impose constraints only
on the \emph{abundance}. That is, $\Psi\left(\mtbfA\right)$ is effective
and $\Phi\left(\mtbfM\right)=0$; the updating rule for the \emph{endmember}
is given in\ \eqref{eq:NMF_updating_M=000026A} by default. Specifically,
the sparse constraints\ \cite{Hoyer_02_NNSP_NMF_l1,Qian_11_TGRS_NMF+l1/2,fyzhu_2014_IJPRS_SSNMF,fyzhu_2014_JSTSP_RRLbS,fyzhu_2014_TIP_DgS_NMF,xiaoqiangLu_2013_TGRS_ManifoldSparseNMF}
and the spatial (like, manifold, graph) constraint\ \cite{Cai_11_PAMI_GNMF,JmLiu_12_SlectedTopics_W-NMF,fyzhu_2014_IJPRS_SSNMF,xiaoqiangLu_2013_TGRS_ManifoldSparseNMF}
are the most popular ones.

\subsubsection{W-NMF\ \cite{JmLiu_12_SlectedTopics_W-NMF} (or G-NMF\ \cite{Cai_11_PAMI_GNMF})}

The local neighborhood weight regularized NMF (W-NMF)~\cite{JmLiu_12_SlectedTopics_W-NMF}
assumes that hyperspectral pixels are distributed on a manifold; the
authors exploit appropriate weights in the local neighborhood to enhance
the spectral unmixing. Specifically, W-NMF employs both the spectral
and spatial information to construct the weight matrix $\mbox{\ensuremath{\mtbfW}}\in\mtbbR_{+}^{N\times N}$\ \cite{Prasad_2014_SPM_ManifoldFeatureExtraction},
where $W_{ij}$ is the similarity between the pixel $i$ and $j$.
While, G-NMF\ \cite{Cai_11_PAMI_GNMF} is a general NMF extension,
which only employs the pixel feature to form the graph.

In W-NMF (or G-NMF), the constraint on the \emph{abundance} is $\Psi\left(\mtbfA\right)=\mttrace\left(\mtbfA\mtbfL\mtbfA^{\intercal}\right)$,
where $\mtbldL=\mtbldD-\mtbldW$ is the graph Laplacian learned from
the hyperspectral image, and $\mtbfD$ is the degree matrix whose
diagonal elements are column sums of $\mtbfW$. The updating rule
of\emph{ abundances} for the W-NMF is
\begin{equation}
A_{kn}\leftarrow\frac{A_{kn}\left(\mtbfM^{\top}\mtbfY+\lambda\mtbfA\mtbfW\right)_{kn}}{\left(\mtbfM^{\top}\mtbfM\mtbfA+\lambda\mtbfA\mtbfD\right)_{kn}},\label{eq:W-NMF_updating_A}
\end{equation}
where $\lambda$ is a balancing parameter for the Graph constraint.

\subsubsection{Typical Sparsity constrained NMF (e.g., $\ell_{1}$-NMF\ \cite{Hoyer_02_NNSP_NMF_l1}
and $\ell_{1/2}$-NMF\ \cite{Qian_11_TGRS_NMF+l1/2})}

The sparsity constrained NMFs\ \cite{Hoyer_02_NNSP_NMF_l1,Qian_11_TGRS_NMF+l1/2,yingWang_2015_TIP_RobustUnmixing,fyzhu_2014_IJPRS_SSNMF,fyzhu_2014_TIP_DgS_NMF,fyzhu_2014_JSTSP_RRLbS}
are the most successful methods for the HU task. Those methods assume
that most hyperspectral pixels are mixed with parts of (not all) \emph{endmembers},
and exploits all kinds of sparse constraints on the \emph{abundance}.
The $\ell_{1}$-norm is one of the most benchmark sparsity constraint.
Considering it gives the $\ell_{1}$-NMF\ \cite{Hoyer_02_NNSP_NMF_l1},
where the constraint is $\Psi\left(\mtbfA\right)=\left\Vert \mtbfA\right\Vert _{1}=\sum_{k,n}\left|A_{kn}\right|$.
The updating rule for the \emph{abundance} is
\begin{equation}
A_{kn}\leftarrow\frac{A_{kn}\left(\mtbfM^{\top}\mtbfY\right)_{kn}}{\left(\mtbfM^{\top}\mtbfM\mtbfA+\lambda\right)_{kn}}.\label{eq:L1-NMF_updating_A}
\end{equation}
Although the $\ell_{1}$-NMF makes sense, the lasso constraint~\cite{Tibshirani_94_Statist_Lasso,Donoho_96_ITIT_CS}
could not enforce further sparse when the full additivity constraint
is used, limiting the effectiveness~\cite{Qian_11_TGRS_NMF+l1/2}.
Accordingly, Dr. Qian proposed the state-of-the-art $\ell_{1/2}$-norm
constrained NMF, where $\Psi\left(\mtbfA\right)=\left\Vert \mtbfA\right\Vert _{1/2}\approx\sum_{k,n}\left|A_{kn}+\xi\right|^{1/2}$,
$\xi$ is a small positive value to ensure the numerical condition.
The updating rule for the \emph{abundance} is given as
\begin{equation}
A_{kn}\leftarrow\frac{A_{kn}\left(\mtbfM^{\top}\mtbfY\right)_{lk}}{\left(\mtbfM^{\top}\mtbfM\mtbfA+0.5\lambda\left(\mtbfA+\xi\right)^{-1/2}\right)_{lk}}.\label{eq:L1/2-NMF_updating_A}
\end{equation}
where $\lambda$ in Eqs.\ \eqref{eq:L1-NMF_updating_A} and\ \eqref{eq:L1/2-NMF_updating_A}
are the balancing parameter for the sparsity constraints\ \cite{Hoyer_02_NNSP_NMF_l1,Qian_11_TGRS_NMF+l1/2}.

\subsubsection{DgS-NMF\ \cite{fyzhu_2014_TIP_DgS_NMF} and RRLbS\ \cite{fyzhu_2014_JSTSP_RRLbS}}

To reduce the solution space, the state-of-the-art HU methods exploit
various constraints on the \emph{abundances} and on the \emph{endmembers}.
However, they generally employ an identical strength of constraints
on all the factors, which may not meet the practical situation. Instead,
Dr. Zhu\ \cite{fyzhu_2014_TIP_DgS_NMF} observed that the mixed level
of each pixel varies over image grids. Based on this prior, he proposed
a novel method to learn a data-guided map (DgMap, i.e., $\mtbfh\in\mtbbR_{+}^{N}$),
which aims to describe the mixed level of each pixel. Through this
DgMap, the $\ell_{p}\left(0<p<1\right)$ constraint is applied in
an adaptive manner. For each pixel, the choice of $p$ is tightly
related to the corresponding value in the DgMap. 

In DgS-NMF, the sparsity constraint on the \emph{abundance} is $\Psi\left(\mtbfA\right)=\sum_{k,n}\left(A_{kn}\!+\!\xi\right)^{1\!-H_{kn}}$,
where $\mtbfH=\mtbfone_{K}\mtbfh^{\intercal}\in\mtbbR_{+}^{K\times N}$,
$\mtbfh\in\mtbbR_{+}^{N}$ is the DgMap, and $\xi$ is a small positive
value to ensure numerical conditions. The updating rule for the \emph{abundance}
is 
\begin{equation}
A_{kn}\leftarrow\frac{A_{kn}\left(\mtbfM^{\top}\mtbfY\right)_{kn}}{\left(\mtbfM^{\top}\mtbfM\mtbfA+\lambda\left(\mtbfone-\mtbfH\right)\circ\left(\mtbfA+\xi\right)^{-\mtbfH}\right)_{kn}},\label{eq:DgS-NMF_updating_A}
\end{equation}
where $\circ$ is the Hadamard product between matrices. Note that
the $\ell_{1}$-NMF and $\ell_{1/2}$-NMF are special cases of the
DgS-NMF. Given a constant DgMap with all elements equal to zero, the
DgS-NMF turns into the $\ell_{1}$-NMF. Whereas if each element in
the DgMap is equal to $1/2$, the DgMap guided sparsity constraint
turns into the $\ell_{1/2}$ norm based sparse regularization. 

DgS-NMF\ \cite{fyzhu_2014_TIP_DgS_NMF} is an interesting method.
However, a heuristic algorithm is proposed in\ \cite{fyzhu_2014_TIP_DgS_NMF}
to learn the DgMap, which is ineffective for the vast smooth areas
in the image.  It is expected that the more accurate DgMap constraint
would bias the solution to the more satisfactory local minima. Besides,
the state-of-the-art method generally ignores the badly degraded (i.e.,
outlier) channels in the hyperspectral image. To address the above
two problems, a \textbf{r}obust \textbf{r}epresentation and \textbf{l}earning-\textbf{b}ased
\textbf{s}parsity (RRLbS) method is proposed in\ \cite{fyzhu_2014_JSTSP_RRLbS}
by emphasizing both robust representation and learning-based sparsity.
Specifically, the $\mtloss\left\{ \cdot,\cdot\right\} $ in\ \eqref{eq:generalized_HU_loss}
is set as the $\ell_{2,1}$-norm based loss. The new loss is better
at preventing the outlier channels from dominating the objective.
The constraint in RRLbS is $\Psi\left(\mtbfA\right)=\sum_{k,n}\left(A_{kn}+\xi\right)^{1-H_{kn}}$,
which is similar to the DgS-NMF. However, due to the robust loss function
and the simultaneous learning process for DgMaps, the updating rule
in RRLbS is quite different from that of DgS-NMF: 
\begin{align}
M_{lk} & \leftarrow\frac{M_{lk}\left(\mtbfU\mtbfX\mtbfA^{\top}\right)_{lk}}{\left(\mtbfU\mtbfM\mtbfA\mtbfA^{\top}\right)_{lk}},\label{eq:RRLbS_updating_M}\\
A_{kn} & \leftarrow\frac{A_{kn}\left(\mtbfM^{\top}\mtbfU\mtbfX\right)_{kn}}{\left(\mtbfM^{\top}\mtbfU\mtbfM\mtbfA+\lambda\left(1-\mtbfH\right)\circ\left(\mtbfA+\xi\right)^{-\mtbfH}\right)_{kn}},\label{eq:RRLbS_updating_A}
\end{align}
where $\mtbfU\in\mtbbR_{+}^{L\times L}$ is a nonnegative diagonal
matrix\footnote{The $l^{\mtth}$ diagonal entry $U_{ll}$ is set as $U_{ll}\!=\!1/\left(2\sqrt{\bigl\Vert\left(\mtbfM\mtbfA-\mtbfX\right)^{l}\bigr\Vert_{2}^{2}+\epsilon}\right)$,
where $\epsilon$ is typically set $10^{-8}$ to avoid singular failures.}. The most direct clue to estimate DgMap $\mtbfh\in\mtbbR_{+}^{N}$
is its crucial dependence upon \emph{abundances.} Once getting the
stable \emph{abundance $\mtbfA$}, $\mtbfh=\left\{ h_{n}\right\} _{n=1}^{N}$
could be efficiently estimated as follows: 
\begin{equation}
\mtbfH=\mtbfone_{K}\mtbfh^{\top},h_{n}=S\left(\mtbfa_{n}\right),\ \forall n\in\left\{ 1,2,\cdots,N\right\} \label{eq:RRLbS_updating_H}
\end{equation}
where $S\left(\mtbfa_{n}\right)$ is the Gini index\ \cite{Hurley_2009_TIT_ComparingSparseMeasures,fyzhu_2014_JSTSP_RRLbS}
that measures the sparsity of column vectors in $\mtbfA$. The above
updating process for $\left\{ \mtbfM,\mtbfA\right\} $  and $\mtbfh$
 is iterated, which is expected to generate a sequence of ever improved
estimates until convergences.

\subsubsection{Structure (or Graph) and Sparsity constrained NMF (SS-NMF\ \cite{fyzhu_2014_IJPRS_SSNMF}
and GL-NMF\ \cite{xiaoqiangLu_2013_TGRS_ManifoldSparseNMF})}

There are two methods\ \cite{fyzhu_2014_IJPRS_SSNMF,xiaoqiangLu_2013_TGRS_ManifoldSparseNMF}
considering both the spatial (like graph) constraint and the sparsity
constraint. In the Stucutred sparsed NMF (SS-NMF), the constraint
is $\Psi\left(\mtbfA\right)=\mttrace\left(\mtbfA\mtbfL\mtbfA^{\top}\right)+\frac{\alpha}{\lambda}\left\Vert \mtbfA\right\Vert _{1}$,
where the graph Laplacian $\mtbfL$ is learned via a novel method
that considers both the spectral and spatial information in the hyperspectral
image. The updating rule for the \emph{abundance} is as follows:
\begin{equation}
A_{kn}\leftarrow\frac{A_{kn}\left(\mtbldM^{\top}\mtbfX+\lambda\mtbldA\mtbldW\right)_{kn}}{\left(\mtbldM^{\top}\mtbldM\mtbldA+\lambda\mtbldA\mtbldD+\alpha\right)_{kn}}.\label{eq:SS-NMF_updating_A}
\end{equation}

In GL-NMFE\ \cite{xiaoqiangLu_2013_TGRS_ManifoldSparseNMF}, the
constraint is $\Psi\left(\mtbfA\right)=\mttrace\left(\mtbfA\mtbfL\mtbfA^{\top}\right)+\frac{\alpha}{\lambda}\left\Vert \mtbfA\right\Vert _{1/2}$,
where the authors ignore the spatial information inherent in the hyperspectral
image and only employ the spectral information to construct the graph
Laplacian $\mtbfL$. The updating rule for the \emph{abundance} is:
\begin{equation}
A_{kn}\leftarrow\frac{A_{kn}\left(\mtbldM^{\top}\mtbfX+\lambda\mtbldA\mtbldW\right)_{kn}}{\left(\mtbldM^{\top}\mtbldM\mtbldA+\lambda\mtbldA\mtbldD+1/2\alpha\mtbfA^{-1/2}\right)_{kn}}.\label{eq:GL-NMF_updating_A}
\end{equation}

\subsubsection{Correntropy based NMF (CE-NMF)\ \cite{yingWang_2015_TIP_RobustUnmixing}}

It is well known that the Euclidean loss is prone to outliers\ \cite{fpNie_2010_NIPS_JointL21_featureSelection,huaWang_2014_ICML_RobustMetricLearning}.
Accordingly, Dr. Wang employed the correntropy metric to measure
the reconstruction error; the $\ell_{1}$ norm based sparse constraint
is considered, resulting in the new robust objective 
\begin{align}
\min_{\mtbfM\geq\mtbfzero,\mtbfA\geq\mtbfzero} & \sum_{l=1}^{L}\left[\!\!-\exp\!\left(\!\!-\frac{\left\Vert \mtbfx^{l}\!-\!\left(\mtbfM\mtbfA\right)^{l}\right\Vert _{2}^{2}}{\sigma^{2}}\right)\!\!\right]\!+\!2\lambda\left\Vert \mtbfA\right\Vert _{1}\!.\label{eq:CE-NMF_ojective}
\end{align}
A half-quadratic optimization technique is proposed to convert the
complex optimization problem\ \eqref{eq:CE-NMF_ojective} into an
iteratively reweighted NMF problem. As a result, the optimization
can adaptively assign small weights to noisy channels and emphasize
on noise-free channels. The updating rule is 
\begin{equation}
\widehat{M}_{lk}\leftarrow\frac{\widehat{M}_{lk}\left(\mhbfX\mtbfA^{\intercal}\right)_{lk}}{\left(\mhbfM\mtbfA\mtbfA^{\intercal}\right)_{lk}},\quad A_{kn}\leftarrow\frac{A_{kn}\left(\mhbfM^{\intercal}\mhbfX\right)_{kn}}{\left(\mhbfM^{\intercal}\mhbfM\mtbfA+\lambda\right)_{kn}},\label{eq:CE-NMF_updating_M=000026A}
\end{equation}
where $\mhbfX\!=\!\mtbfU^{1/2}\mtbfX$, $\mhbfM\!=\!\mtbfU^{1/2}\mtbfM$
and $\mtbfU\in\mtbbR_{+}^{L\times L}$ is a diagonal matrix with the
$l^{\mtth}$ element as $U_{ll}=\exp\left(-\frac{\left\Vert \mtbfx^{l}-\left(\mtbfM\mtbfA\right)^{l}\right\Vert _{2}^{2}}{\sigma^{2}}\right)$.

\subsection{The NMF extensions with constraints on the endmembers \label{sub:NMF_Constrained_endmembers}}

We review the NMF extensions that impose constraints only on the \emph{endmember}.
That is, $\Phi\left(\mtbfM\right)$ is effective and $\Psi\left(\mtbfA\right)=0$

\subsubsection{Minimum Volume Constrained NMF (MVC-NMF)\ \cite{Miao_07_ITGRS_NMFMVC}}

The MVC-NMF combines the property of both the geometric methods and
statistical methods. It aims to find the \emph{endmembers}, which
compose the minimum volume simplex that circumscribes the hyperspectral
data scatters. Given the \emph{endmembers}, the simplex volume is
defined as
\begin{equation}
\Phi\left(\mtbfM\right)=\frac{1}{\left(K-1\right)!}\left|\mtdet\left(\left[\mtbfm_{2}\!-\!\mtbfm_{1},\cdots,\mtbfm_{K}\!-\!\mtbfm_{1}\right]\right)\right|.\label{eq:MVC-NMF_constraint}
\end{equation}
The standard Euclidean is used to measure the representation error.
For the sake of easy optimization, the objective function is finalized
as $\min_{\mtbfM,\mtbfA}f\left(\mtbfM,\mtbfA\right)$ where 
\[
f\left(\mtbfM,\mtbfA\right)=\frac{1}{2}\left\Vert \mtbfX-\mtbfM\mtbfA\right\Vert _{F}^{2}+\frac{\lambda}{2\left(K-1\right)!}\mtdet^{2}\left(\!\left[\mtbfone_{K},\mhbfM^{\top}\right]^{\top}\!\right)
\]
where $\mhbfM$ is a low dimensional transform of $\mtbfM$, i.e.,
$\mhbfM=\mtbfP^{\intercal}\left(\mtbfM-{\bf \bm{\mu}\mtbfone}_{K}^{\intercal}\right)$;
$\mtbfP\in\mtbbR^{L\times\left(K-1\right)}$ is formed by the $K-1$
most significant principle components of hyperspectral data $\mtbfX$
and $\bm{\mu}$ is the mean of data $\mtbfX$. The updating rule is
based on the projected gradient algorithm, which is given as follows:
\begin{alignat}{1}
\mtbfM^{\left(t+1\right)} & =\max\left\{ \mtbfzero,\ \mtbfM^{\left(t\right)}-\alpha^{\left(t\right)}\nabla_{\mtbfM}f\left(\mtbfM^{\left(t\right)},\mtbfA^{\left(t\right)}\right)\right\} \label{eq:MVC-NMF_updating_M}\\
\mtbfA^{\left(t+1\right)} & =\max\left\{ \!\mtbfzero,\ \mtbfA^{\left(t\right)}-\beta^{\left(t\right)}\nabla_{\mtbfA}f\left(\mtbfM^{\left(t+1\right)},\mtbfA^{\left(t\right)}\right)\!\right\} \label{eq:MVC-NMF_updating_A}
\end{alignat}
where $\alpha^{\left(t\right)}$ and $\beta^{\left(t\right)}$ are
the learning rates. They can be fixed at small values or determined
by the Armijo rule\ \cite{bertsekas2014constrained,Miao_07_ITGRS_NMFMVC}.

\subsubsection{Endmember Dissimilarity Constrained NMF (EDC-NMF)\ \cite{nWang_13_SelectedTopics_EDC-NMF}}

Inspired by the MVC-NMF\ \cite{Miao_07_ITGRS_NMFMVC}, Dr. Wan proposed
the \emph{Endmember} Dissimilarity Constrained NMF (EDC-NMF). The
core assumption is that due to the high spectral resolution of hyperspectral
sensors, the \emph{endmember} spectra should be smooth itself and
different as much as possible from each other. The EDC constraint
on the \emph{endmember} is 
\begin{equation}
\Phi\left(\mtbfM\right)=\sum_{i=1}^{K-1}\sum_{j=i+1}^{K}\left\Vert \nabla\mtbfm_{i}-\nabla\mtbfm_{j}\right\Vert _{2}^{2},
\end{equation}
where $\nabla\mtbfm_{i}$=$\left[m_{2,i}-m_{1,i},\cdots,m_{L,i}-m_{L-1,i}\right]$
is the gradient along the spectral dimension.

The multiplicative update rule is used to optimize the objective function,
resulting in the algorithms for the \emph{abundances} in\ \eqref{eq:NMF_updating_M=000026A}
and for the \emph{endmembers} as follows\emph{ }
\begin{equation}
M_{lk}\leftarrow\frac{M_{lk}\left(\mtbfX\mtbfA^{\top}-0.5\lambda\nabla_{\mtbfM}\Phi\left(\mtbfM\right)\right)_{lk}}{\left(\mtbfM\mtbfA\mtbfA^{\top}\right)_{lk}},\label{eq:EDC-NMF_updating_M}
\end{equation}
where $\nabla_{\mtbfM}\Phi\left(\mtbfM\right)=2\mtbfH\mtbfH^{\top}\mtbfM\mtbfT$;
$\mtbfT$ is a specially designed matrix, i.e., $\mtbfT=K\mtbfI_{K}-\mtbfone_{K\times K},$
$\mtbfI_{K}$ is a $K\times K$ identical matrix and $\mtbfone_{K\times K}$
is a matrix whose all elements are $1$; $\mtbfH=\left[-\mtbfp^{\top};\mtbfI_{L-1}\right]+\left[-\mtbfI_{L-1};\mtbfq^{\top}\right]$,
$\mtbfI_{L-1}$ is a $\left(L-1\right)\times\left(L-1\right)$ identical
matrix, $\mtbfp=\left[1,0,\cdots0\right]^{\top}\in\mtbbR^{L-1}$ and
$\mtbfq=\left[0,\cdots,0,1\right]^{\top}\in\mtbbR^{L-1}$. 

The derivative over \emph{endmembers}, i.e., $\nabla_{\mtbfM}\Phi\left(\mtbfM\right)$,
introduces negative values to the updating rule\ \eqref{eq:EDC-NMF_updating_M}.
To make up this problem, the negative elements in the \emph{endmember}
matrix are required to project to a given nonnegative value after
each iteration. Consequently, the regularization parameter $\lambda$
couldn't be chosen freely, limiting the efficacy of EDC-NMF.

\section{How to Label Ground Truths in the HU?\label{sec:the method to label GT 4 HU}}

In the HU, the ground truth labeling consists of two parts: (1) the
\emph{endmember} labeling, and (2) the \emph{abundance} labeling.
Both parts are challenging. There is only one paper\ \cite{SJia_07_TGRS_SSCBSS}
briefly introducing the ground truth labeling on the HYDICE urban
image. As a result, this is the \textbf{first} paper to provide a
general method to label the hyperspectral images for the HU task.

\subsection{The Endmember labeling\ \label{sub:The-Endmember-Labelling}}

The \emph{endmember} is the ``pure'' material in the image\ \cite{Dias_12_AEORS_HUoverview}.
What is the ``pure'' material? From the Chemistry perspective, the
``pure'' material (i.e.\emph{ endmember}) can be the pure substance.
However, this ideal assumption makes the HU impossible to solve---there
can be thousands of ``pure'' substance in the image scene. Such case
leads to much more unknown variables than the observation itself in
the hyperspectral image, which is obviously a severely underconstrained
problem\ \cite{ALevin_2008_PAMI_closedFormd}. 

Alternatively, it is widely accepted that the notion of ``pure material''
 is subjective and problem dependent\ \cite{Dias_12_AEORS_HUoverview},
which is similar with the determination of classes in the hyperspectral
classificiation task. Such setting requires lots of professional analyses
and understandings of the image scene\ \cite{landgrebe1998multispectral}.
Taking the Cuprite image for example, it may take Earth scientists
several weeks to analyze all the \emph{endmembers} in the scene. The
good news is that three factors make it easy to determine \emph{endmembers}:
\textbf{(1)} there are many HU papers publishing the \emph{endmember}
information of the common hyperspectral images, such as the HYDICE
urban\ \cite{Qian_11_TGRS_NMF+l1/2,SJia_07_TGRS_SSCBSS,Jia_09_TGRS_ConstainedNMF},
Cuprite\ \cite{Jose_05_TGRS_Vca,Qian_11_TGRS_NMF+l1/2,xiaoqiangLu_2013_TGRS_ManifoldSparseNMF},
Washington DC Mall\ \cite{nWang_13_SelectedTopics_EDC-NMF}, Moffett
Field\ \cite{Dobigeon_ISP_BaysianHU,Dobigeon_TIP_SemiSupervised_LSU,DBLP:journals/tgrs/YokoyaCI14}
etc; \textbf{(2)} for the images whose scenes are simple, it is easy
to determine the \emph{endmembers}, e.g., the Samson\ \cite{fyzhu_2014_IJPRS_SSNMF},
Japser Ridge\ \cite{fyzhu_2014_TIP_DgS_NMF,fyzhu_2014_JSTSP_RRLbS}
and San Deigo Airport\ \cite{gillis2013sparse} etc. \textbf{(3)}
the existing methods, like virtual dimensionality (VD)\ \cite{Chang_2004_TGRS_VD},
is also helpful to determine the number of \emph{endmembers} in the
hyperspectral image.

After the determination of \emph{endmembers}, we need their spectral
signatures, which could be obtained as below:

\textbf{1)}. \uline{They can be manually chosen from the hyperspectral
image}. In some images, the ``pure'' pixels have been explicitly marked
in the published papers. For example, the \emph{endmembers} for the
Washington DC Mall image are explicitly marked in Fig. 13 in\ \cite{nWang_13_SelectedTopics_EDC-NMF}.
In the HYDICE urban, the location of the \emph{endmember }pixel for
``asphalt'' is provided in the paper\ \cite{SJia_07_TGRS_SSCBSS},
which is then compared with the asphalt spectrum in the standard spectral
library. In other hyperpsectral images, like the Jasper Ridge, Samson,
etc., we may have to select the \emph{endmembers} from the image based
on our understandings. 

\textbf{2)}. The \emph{endmember} signature can be found in the \textbf{USGS}
mineral spectral library. The Cuprite hyperspectral image is one of
the typical hyperspectral image\ \cite{nWang_13_SelectedTopics_EDC-NMF}
in this type.

\subsection{The Abundance labeling\ \label{sub:The-Abundance-Labelling}}

The \emph{abundance} provides the composite percentage of all \emph{endmembers}
at each pixel (cf. Section\ \ref{sec:Review_NMF=000026Extensions}).
Since the \emph{abundance} is continuous between 0 and 1, it is impossible
to manually label the \emph{abundances}---Human can not tell the difference
between 30\% and 40\% of an \emph{endmember} at a pixel; besides,
it is too tedious and labor intensive to label all pixels. Instead,
we propose a learning method to label the \emph{abundance}. 

Given the determined $K$ \emph{endmembers}, the \emph{abundance}
labeling becomes a least square problem with the nonnegative and full
additivity constraints, which is indeed the supervised unmixing problem
as reviewd in Section\ \ref{sec:3_types_methods}. Such algorithm
is far more easier than the unsupervised unmixing methods. The general
quadratic programming algorithm with constraints (like the \emph{quadprog}
in Matlab) is helpful to estimate the \emph{abundances}. Moreover,
we may make use of the state-of-the-art unmxing methods reviewed in
Section\ \ref{sub:Nonnegative-Matrix-Factorization} and\ \ref{sub:NMF_Constrained_Abundance}
by treating the \emph{endmembers} as a fixed input rather than unknown
variables and by only updating the \emph{abundances}. The state-of-the-art
unmxing methods are promising to achieve very good ground truths because
of the different priors on the \emph{abundance}.

\subsection{How to verify our labeling ground truth for the HU study?\label{sub:verify_labelling_result}}

What is a good ground truth for the given hyperspectral image? To
verify it, we come up with two evaluation methods:

\textbf{1)}. \uline{compare our labeling result with the ground
truth posted on the published papers}. It is necessary to ensure the
high similarity between them. Note that, only the illustration of
ground truths (not the ground truth itself) is posted in the paper.
Accordingly, we can only compare them visually, not quantitatively.
For example, we may compare the shape of the \emph{endmember} signatures
and the illustration of \emph{abundances}. 

\textbf{2)}. check the labeling result based on our understanding
of the hyperspectral image---the labeling result should be consistent
with our understanding of the image.

\textbf{3)}. it is reasonable to assume that similar pixels have similar
\emph{abundances}\ \cite{fyzhu_2014_IJPRS_SSNMF,xiaoqiangLu_2013_TGRS_ManifoldSparseNMF}.
Based on this assumption, we will click many pixels to evaluate their
signature's similarities, and then to check their \emph{abundance's}
similarity. Those two kinds of similarities should be consistent as
well.

\textbf{4)}. for the hyperspectral images, like Cuprite, the ground
truth \emph{endmembers} are selected from the\emph{ }standard spectral
library\ \cite{fyzhu_2014_TIP_DgS_NMF,Qian_11_TGRS_NMF+l1/2,xiaoqiangLu_2013_TGRS_ManifoldSparseNMF,nWang_13_SelectedTopics_EDC-NMF,Jose_05_TGRS_Vca}.
We don't have to verify it. 

We may have to try the different settings on the candidate \emph{endmembers},
on the supervised unmixing algorithms for the \emph{abundance} labeling,
and on the hyper parameters etc (as shown in Algorithm\ \ref{alg:groundTruthLabelling}),
to make a better ground truth.
\begin{algorithm}[t]
\caption{\textbf{to label the ground truth for the HU study\label{alg:groundTruthLabelling}}}
 \textbf{Input} a hyperspectral image $\mtbfX\in\mtbbR_{+}^{L\times N}$,
with $N$ pixels.

\begin{algorithmic}[1] 

\STATE set the \emph{endmembers} and the \emph{endmember} number
$K$.

\REPEAT

\STATE \emph{endmember} labeling via the method in Section\ \ref{sub:The-Endmember-Labelling}.

\STATE \emph{abundance} labeling via the method in Section\ \ref{sub:The-Abundance-Labelling}.

\STATE verify the ground truth labeling.

\UNTIL{meeting the criteria in Section\ \ref{sub:verify_labelling_result}} 

\end{algorithmic} 

\textbf{Output} the ground truth $\mtbfM\in\mtbbR_{+}^{L\times K}$
and $\mtbfA\in\mtbbR_{+}^{K\times N}$.
\end{algorithm}

\subsection{How to transform the benchmark hyperspectral classification (HyC)
datasets for the HU research? \label{sub:HU_labeling_from_classificationResults}}

Compared with the HU datasets, the datasets for the hypserpectral
classification (HyC) task have three obvious advantages. \textbf{(1)}
It is well known that the HyC task has much more benchmark datasets
than the HU task. \textbf{(2)} Different from the HU datasets whose
ground truths are mostly unavailable on the web, the HyC datasets
and their ground truths are mostly available on the web. \textbf{(3)}
the labeling process as well as the label itself for the HyC image
are much easier to understand. Thus, it would be promising to transform
the benchmark HyC datasets for the HU study. However, there is a big
gap between the HyC task and the HU task, which makes the transformation
process challenging. In this paper, we are the first to propose a
promising method to accomplish this transformation task.

Our method consists of three process steps listed as below:

\textbf{1)}. \uline{The number of classes should be consistent
with the number of }\emph{\uline{endmembers}}. Due to the supervised
information involved in the training process, the HyC algorithms are
generally able to handle more classes than the number of \emph{endmembers}
in the HU. Accordingly, we have to combine the similar classes to
reduce the number of \emph{endmembers}.

\textbf{2)}. \uline{Candidate}\emph{\uline{ endmember}}\uline{
labeling}. We need the \emph{endmember} signatures. Given the training
labels for the HyC image, we may select a set of high quality pixels
from the pixel set of each class. The \emph{endmember} signature can
be set as the mean or median signature of all the selected pixels
in that class.

\textbf{3)}. \emph{\uline{Abundance}}\uline{ labeling}. Suppose
we are given a hyperspectral image $\mtbfX\in\mtbbR^{L\times N}$,
with $L$ channels and $N$ samples. The label of all pixels, i.e.,
$\mtbfY\in\mtbbR^{K\times N}$, could be predicted via the state-of-the-art
classification algorithms\ \cite{guangliangCheng_2016_JStars_robustHyperClassification,haichangLi_2016_IJRS_LablePropagationHyperClassification},
where each column $\mtbfy_{n}\in\mtbbR^{K},\forall n\in\left\{ 1,\cdots,N\right\} $
is the 0-1 binary label vector. That is, if pixel $n$ is in the $k^{\mtth}$
class, we define the label vector $\mtbfy_{n}=\left[0,\cdots,0,1,0,\cdots,0\right]^{\top}\in\mtbbR^{K}$
with only the $k^{\mtth}$ element equal to 1, all the others equal
to 0\ \cite{shimingXiang_2012_TNNLS_DLSR}. There are two ways to
treat the candidate \emph{endmember} obtained in the 2nd step. We
may treat them as the final \emph{endmembers}. Accordingly, the \emph{abundance}
labeling becomes the supervised unmixing problem with a special anchoring
constraint as follows:
\begin{align*}
\min_{\mtbfA}\  & \mtloss\left\{ \mtbfX,\widetilde{\mtbfX}\right\} +\lambda\Upsilon\left(\mtbfA\right),\mtsubTo\ \widetilde{\mtbfX}=\mtbfM\mtbfA,\mtbfM\geq\mtbfzero,\mtbfA\geq\mtbfzero,
\end{align*}
where $\Upsilon\left(\mtbfA\right)=\Psi\left(\mtbfA\right)+\frac{\alpha}{\lambda}\left\Vert \mtbfA-\mtbfY\right\Vert _{F}^{2}$
is the constraint on the \emph{abundance}; $\Psi\left(\mtbfA\right)$
is the state-of-the-art sparse/spatial (or both) constraints reviewed
in Section\,\ref{sub:NMF_Constrained_Abundance}; $\left\Vert \mtbfA-\mtbfY\right\Vert _{F}^{2}$
is the anchoring constraint; its aim is to keeps the \emph{abundance}
around the high accuracy classification results.

We may also treat the candidate \emph{endmember}, obtained in the
2nd step, as a good initialization of $\mtbfM$, and update both the
\emph{endmember} $\mtbfM$ and \emph{abundance} $\mtbfA$. The general
model is 
\begin{align*}
\min_{\mtbfM,\mtbfA}\  & \mtloss\left\{ \mtbfX,\widetilde{\mtbfX}\right\} +\lambda\Upsilon\left(\mtbfA\right),\mtsubTo\ \widetilde{\mtbfX}=\mtbfM\mtbfA,\mtbfM\geq\mtbfzero,\mtbfA\geq\mtbfzero,
\end{align*}
where $\Upsilon\left(\mtbfA\right)$ is the \emph{abundance} constraint
reviewed right above. 
\begin{figure}[t]
\noindent \begin{centering}
\subfloat[\textbf{Samson\#1} and its 3 \emph{endmembers}: \#1 Soil, \#2 Tree
and \#3 Water. \label{fig:GT_Samson=0000231}]{\begin{centering}
\includegraphics[width=0.85\columnwidth]{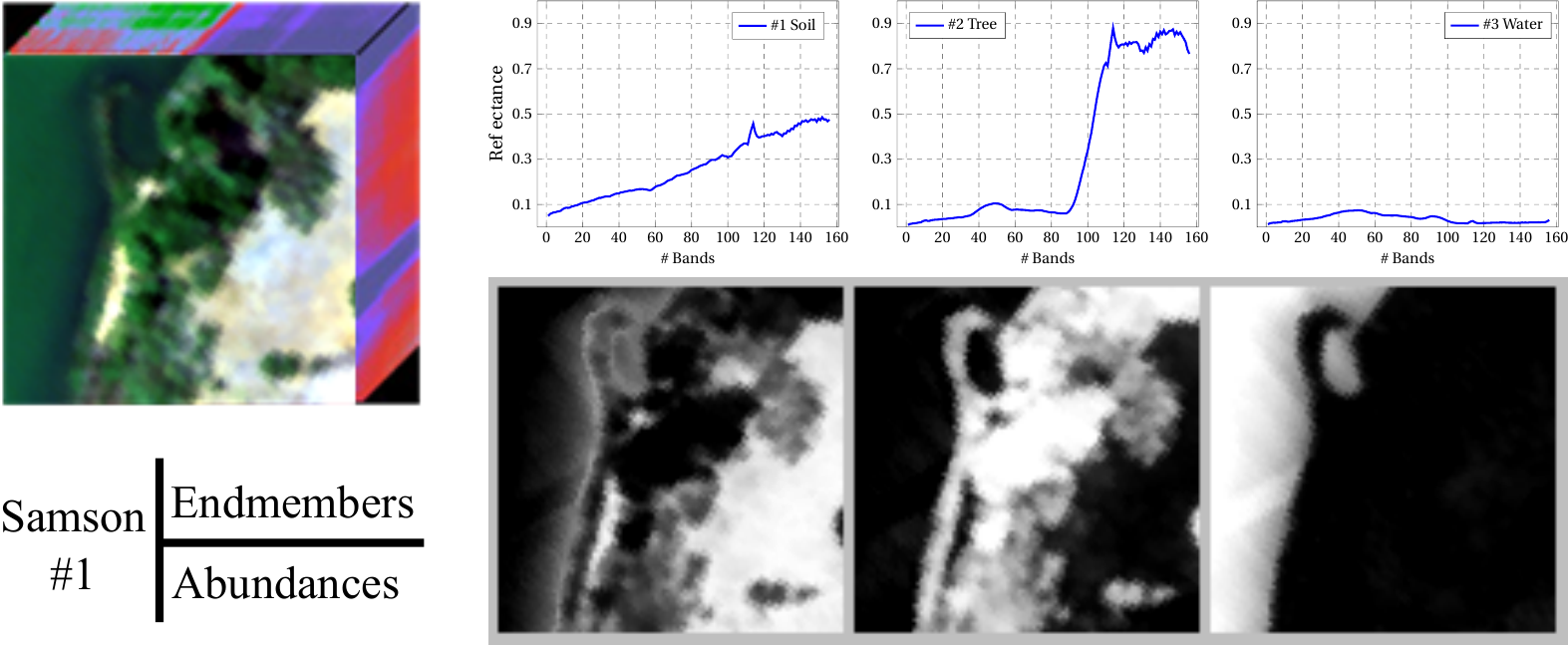}
\par\end{centering}

{\footnotesize{}}{\footnotesize \par}}
\par\end{centering}

\noindent \begin{centering}
\vspace{-0.25cm}
\par\end{centering}

\noindent \begin{centering}
\subfloat[\textbf{Samson\#2} and its 3 \emph{endmembers}: \#1 Soil, \#2 Tree
and \#3 Water.\label{fig:GT_Samson=0000232}]{\begin{centering}
\includegraphics[width=0.85\columnwidth]{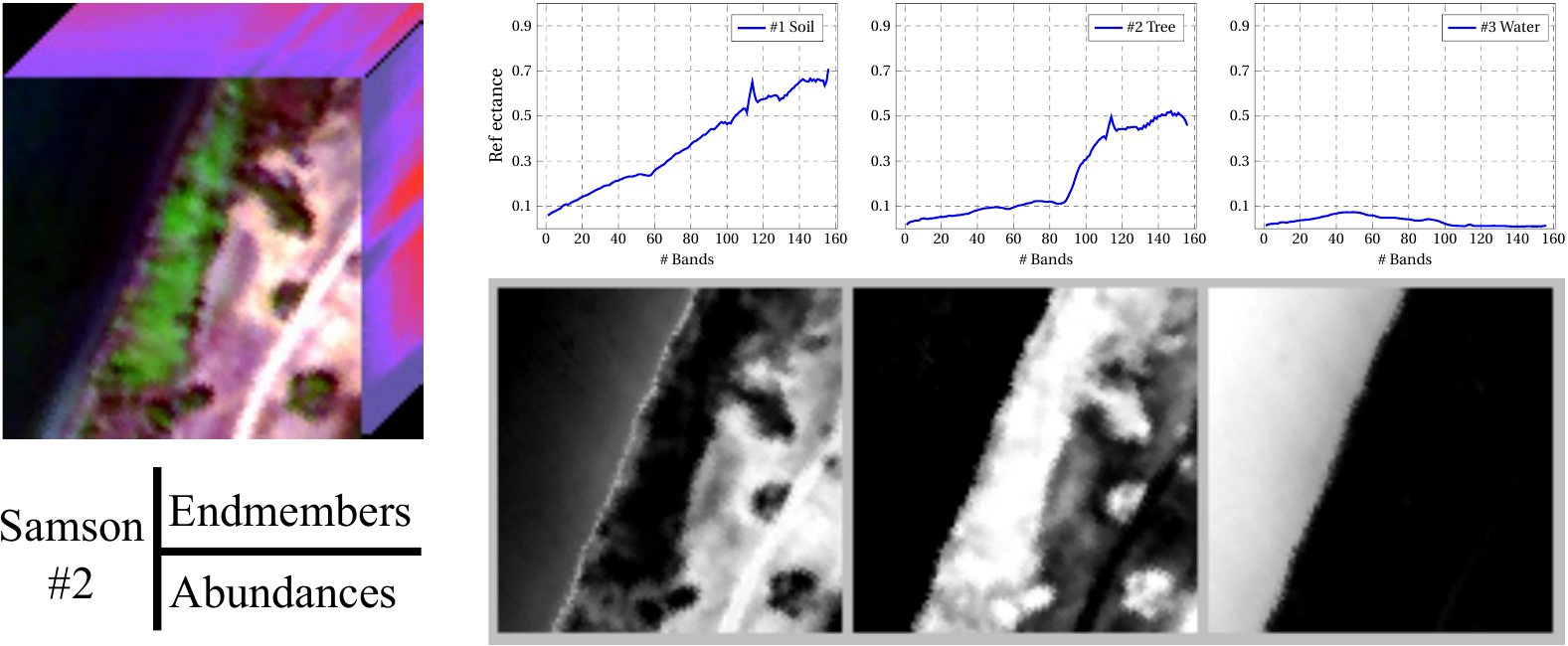}
\par\end{centering}

}
\par\end{centering}

\noindent \begin{centering}
\vspace{-0.25cm}
\par\end{centering}

\noindent \begin{centering}
\subfloat[\textbf{Samson\#3} and its 3 \emph{endmembers}: \#1 Soil, \#2 Tree
and \#3 Water. \label{fig:GT_Samson=0000233}]{\begin{centering}
\includegraphics[width=0.85\columnwidth]{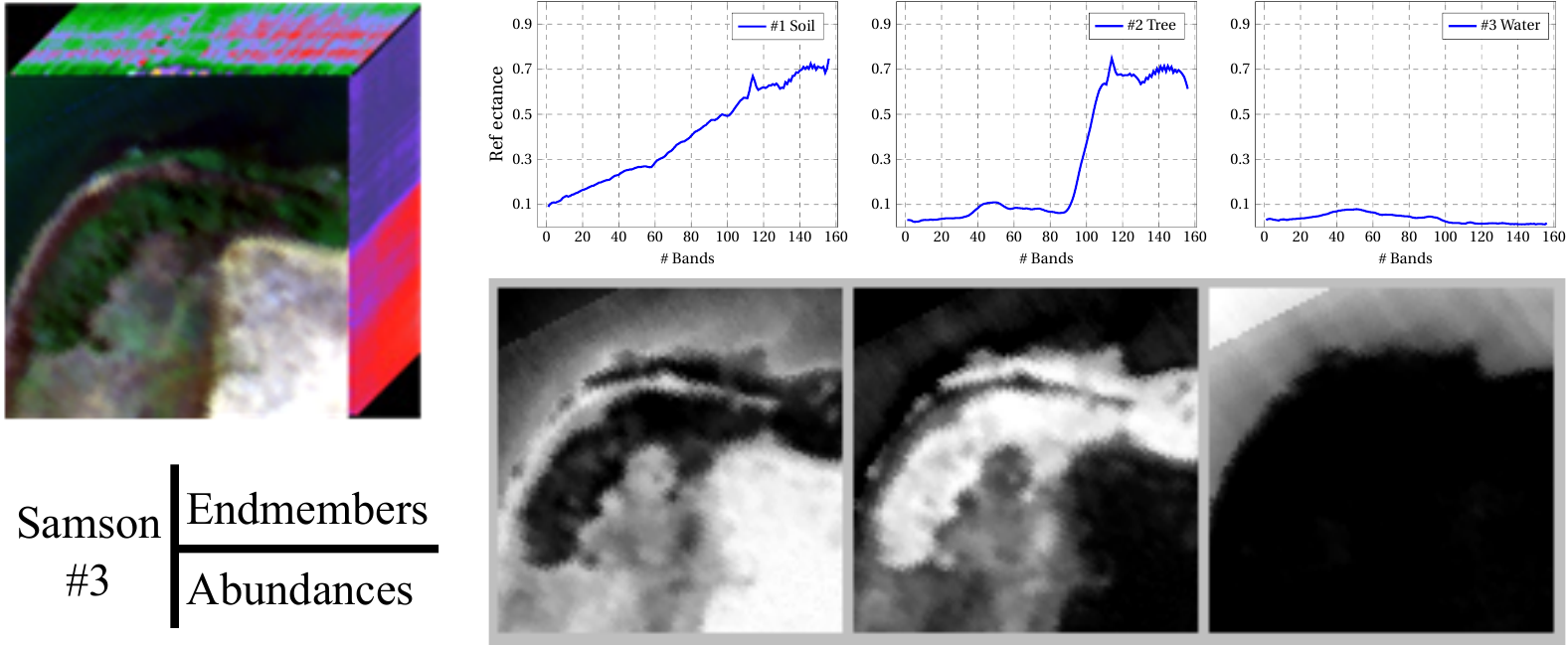}
\par\end{centering}

}
\par\end{centering}

\noindent \begin{centering}
\vspace{-0.25cm}
\par\end{centering}

\caption{The three subimages selected from the Samson hyperspectral image.
They are Samson\#1, Samson\#2 and Samson\#3 respectively, as shown
in Figs\ \ref{fig:GT_Samson=0000231},\,\ref{fig:GT_Samson=0000232}
and\,\ref{fig:GT_Samson=0000233}. Besides, we provide the illustration
of the ground truth \emph{endmembers} and \emph{abundance} maps for
each subimage. \label{fig:GT_3_Samson}}
\end{figure}

\section{Summarization of the 15 Real Hyperspectral Images and their 18 Ground
Truths \label{sec:Real-Hyperspectral-Images}}

\subsection{Samson's three subimages}

The Samson\footnote{downloaded on\ \href{http://opticks.org/confluence/display/opticks/Sample+Data}{http://opticks.org/confluence/display/opticks/Sample+Data}.}
image is illustrated in Fig.\,\ref{fig:7_hyperImages}a, where there
are $952\times952$ pixels\ \cite{fyzhu_2014_IJPRS_SSNMF,fyzhu_2014_JSTSP_RRLbS,fyzhu_2014_TIP_DgS_NMF,fyzhu_2015_PhDthesis}.
Each pixel is observed at 156 bands covering the electromagnetic spectra
from $401\thinspace nm$ to $889\,nm$. As a result, the spectral
resolution is highly up to $3.13\,nm$. This data is in good condition
and not degraded by the blank channels or badly noised channels. The
original image is very large, which could be computationally expensive
for the HU study. Accordingly, three subimages are selected due to
that they contain large transitional areas which consist of lots of
mixed pixels. We name the three ROIs as Samson\#1, Samson\#2 and Samson\#3
respectively for convenience. Please refer to the Fig.\,\ref{fig:7_hyperImages}a
and Table\,\ref{tab:18_realHyperDatasets} for their information,
e.g., their locations, sizes, scenes, number of bands etc. 

All of three ROIs have $95\times95$ pixels. The first ROI (i.e.,
Samson\#1) starts from the $\left(252,\,332\right)$-th pixel in the
original image, cf., Figs.\,\ref{fig:7_hyperImages}a and\,\ref{fig:GT_Samson=0000231}.
The top-left pixel in Samson\#2 corresponds to the $\left(232,\,93\right)$-th
pixel in the original image, cf., Figs.\,\ref{fig:7_hyperImages}a
and\,\ref{fig:GT_Samson=0000232}. Samson\#3's first pixel is the
$\left(643,\,455\right)$-th pixel in the Samson image, cf., Figs.\,\ref{fig:7_hyperImages}a
and\,\ref{fig:GT_Samson=0000233}. Specifically, there are three
\emph{endmembers} in this image, i.e., \textquotedbl{}\#1 Soil\textquotedbl{},
\textquotedbl{}\#2 Tree\textquotedbl{} and \textquotedbl{}\#3 Water\textquotedbl{}
respectively, as shown in Fig.\,\ref{fig:GT_3_Samson}.  
\begin{figure}[t]
\noindent \begin{centering}
\subfloat[\textbf{Jasper Ridge\#1} and its five \emph{endmembers} ``\#1 Road'',
``\#2 Soil'', ``\#3 Water'', ``\#4 Tree'' and ``\#5 Other'' respectively.
\label{fig:GT_JasperRidge=0000231}]{\begin{centering}
\includegraphics[width=0.99\columnwidth]{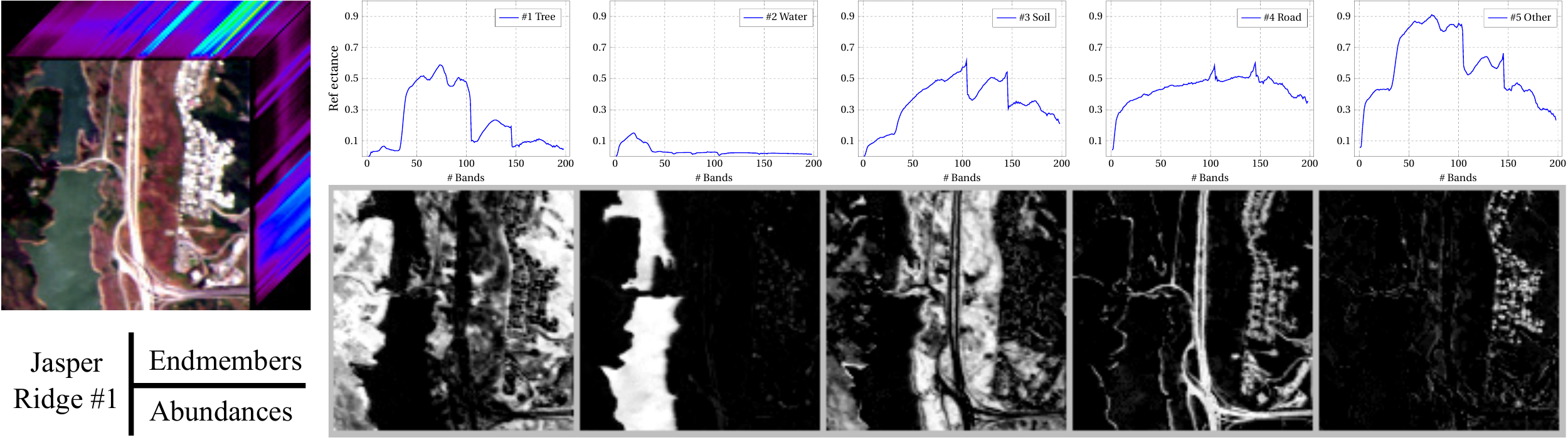}
\par\end{centering}

}\vspace{-0.08cm}
\par\end{centering}

\noindent \begin{centering}
\subfloat[\textbf{Jasper Ridge\#2} and its four \emph{endmembers} ``\#1 Road'',
``\#2 Soil'', ``\#3 Water'' and ``\#4 Tree'' respectively. \label{fig:GT_JasperRidge=0000232}]{\begin{centering}
\includegraphics[width=0.99\columnwidth]{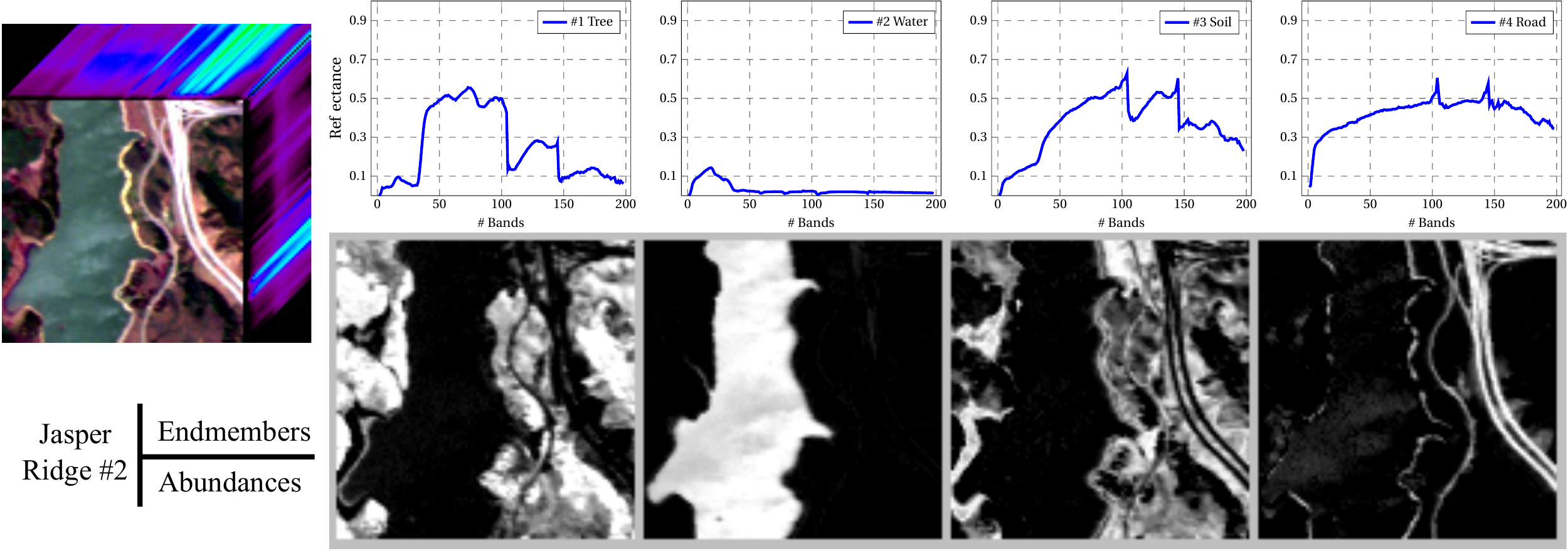}
\par\end{centering}

} \vspace{-0.08cm}
\par\end{centering}

\noindent \begin{centering}
\subfloat[\textbf{Jasper Ridge\#3} and its four \emph{endmembers} ``\#1 Road'',
``\#2 Soil'', ``\#3 Water'' and ``\#4 Tree'' respectively. \label{fig:GT_JasperRidge=0000233}]{\begin{centering}
\includegraphics[width=0.99\columnwidth]{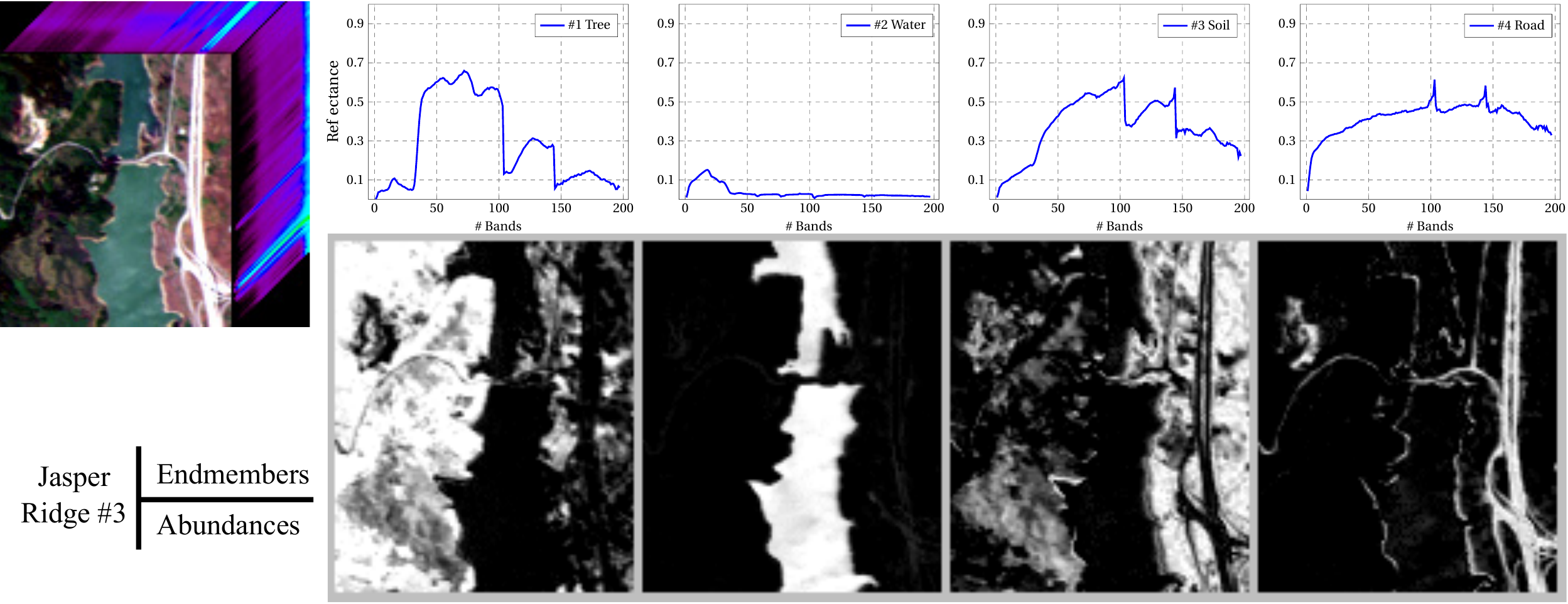}
\par\end{centering}

}
\par\end{centering}

\caption{The three subimages selected from the Jasper Ridge hyperspectral image.
They are Jasper Ridge\#1, Jasper Ridge\#2 and Jasper Ridge\#3 respectively
as shown in Figs\ \ref{fig:GT_JasperRidge=0000231},\,\ref{fig:GT_JasperRidge=0000232}
and\,\ref{fig:GT_JasperRidge=0000233}. Besides, we provide the illustration
of the \emph{endmembers} and the \emph{abundance} maps for each subimage\@.
\label{fig:GT_3_JasperRidge}}
\end{figure}

\subsection{Jasper Ridge's three subimages}

Jasper Ridge is a popular hyperspectral image\,\cite{enviTutorials,rodarmel2002principal},
as shown in Fig.\,\ref{fig:7_hyperImages}b. It was captured via
the AVIRIS (Airborne Visible/Infrared Imaging Spectrometer) sensor
by the Jet Propulsion Laboratory (JPL). The ground sample distance
(GSD) is $20\thinspace m$\ \cite{rodarmel2002principal}. There
are $512\times614$ pixels in it. Each pixel is recorded at 224 electromagnetic
bands ranging from $380\thinspace nm$ to $2,500\thinspace nm$. The
spectral resolution is highly up to $9.46\,nm$. Due to the dense
water vapor and atmospheric effects, we remove the spectral bands
1\textendash 3, 108\textendash 112, 154\textendash 166 and 220\textendash 224,
remaining 198 bands. This is a common preprocessing step if the HU
methods do not focus on the robust learning\ \cite{fyzhu_2014_AAAI_ARSS,yingWang_2015_TIP_RobustUnmixing,fyzhu_2014_JSTSP_RRLbS,guangliangCheng_2016_JStars_robustHyperClassification,xiaoping_2017_ICASSP,haichangLi_2016_IJRS_LablePropagationHyperClassification}.
 
\begin{figure}[t]
\noindent \begin{centering}
\includegraphics[width=0.99\columnwidth]{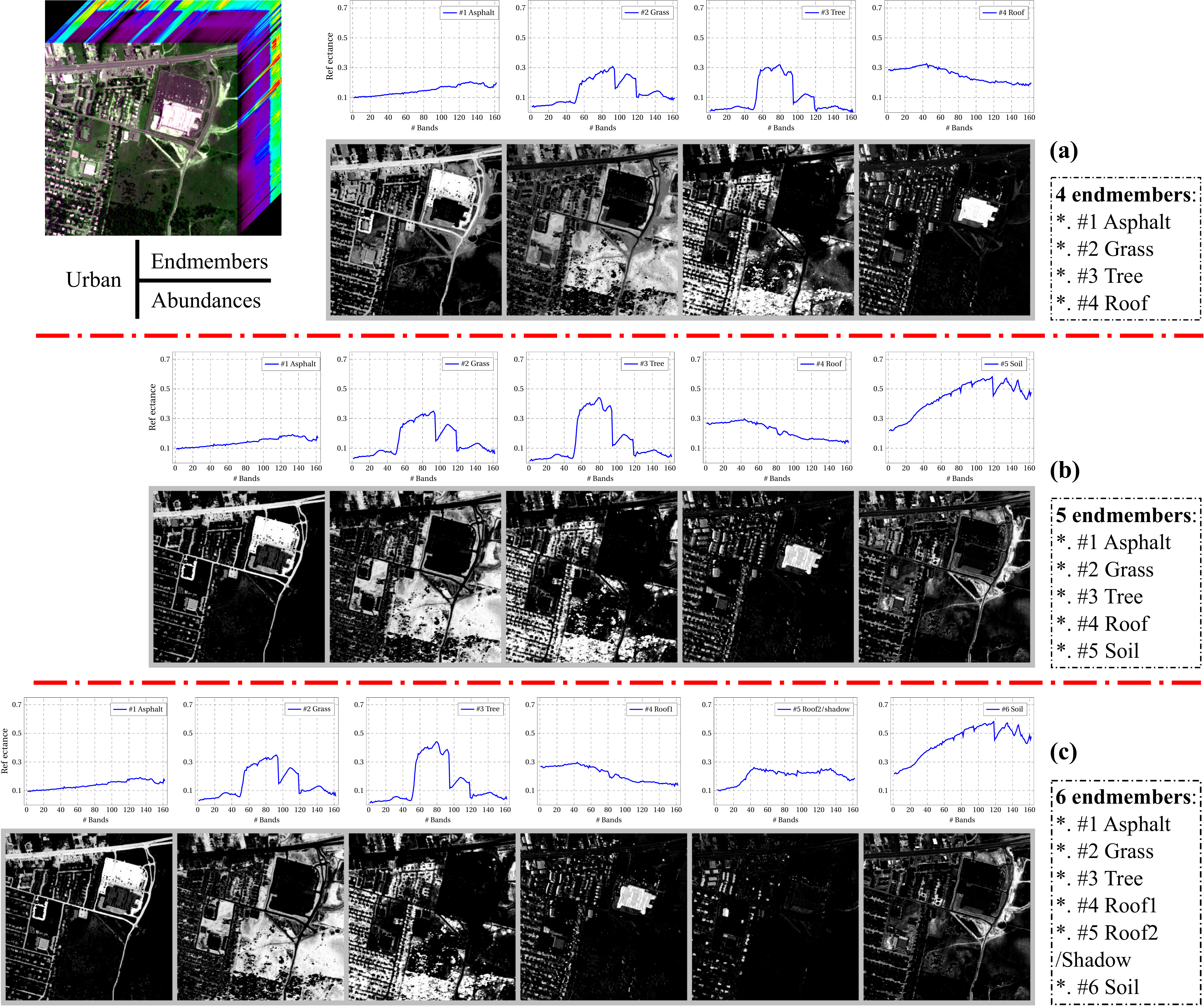}
\par\end{centering}

\caption{The three versions of ground truths for the Urban hyperspectral image.
(a) illustrates the ground truth that has 4 \emph{endmembers}; (b)
shows the ground truth that has 5 \emph{endmembers}; (c) the ground
truth that has 6 \emph{endmembers}. \label{fig:GT_Urban}}
\end{figure}

Since the full hyperspectral image, as shown in Fig.\,\ref{fig:7_hyperImages}b,
is too complex to get the ground truth, we consider three subimages,
named as Jasper Ridge\#1, Jasper Ridge\#2 and Jasper Ridge\#3 respectively.
Please refer to the Fig.\,\ref{fig:7_hyperImages}b and Table\,\ref{tab:18_realHyperDatasets}
for their information. The first subimage (i.e., Jasper Ridge\#1)
consists of $115\times115$ pixels, whose top-left pixel is the $(1,\thinspace272)$-th
pixel in the original image. There are five \emph{endmembers} in Jasper
Ridge\#1, i.e., ``\#1 Road'', ``\#2 Soil'', ``\#3 Water'', ``\#4 Tree''
and ``\#5 Other'', as shown in Fig.\,\ref{fig:GT_JasperRidge=0000231}. 

Jasper Ridge\#2 contains $100\times100$ pixels. The first pixel is
the $\left(105,\,269\right)$-th pixel in the original image. There
are four \emph{endmembers} latent in this subimage: ``\#1 Road'',
``\#2 Soil'', ``\#3 Water'' and ``\#4 Tree'', as shown in Fig.\,\ref{fig:GT_JasperRidge=0000232}.
Since we provided the information of this subimage in the papers\,\cite{fyzhu_2014_TIP_DgS_NMF,fyzhu_2014_IJPRS_SSNMF,fyzhu_2014_JSTSP_RRLbS}
and published the data (including the ground truth) on the homepage\ \cite{fyzhu_2014_GroundTruth4HU}
in 2015, it has been widely used in more than 10 HU papers\ \cite{rodarmel2002principal,vargas2015colored,tong2016nonnegative,shu2015multilayer,tong2017region,vasuki2017clustering,ganesanmaximin,fu2015low,yingWang_2015_TIP_RobustUnmixing,aggarwal2016hyperspectral}
to verify the state-of-the-art algorithms. The third subimage (i.e.,
Jasper Ridge\#3) consists of $122\times104$ pixels. The first (i.e.,
top-left) pixel is the $\left(1,\,246\right)$-th pixel in the original
image. There are four \emph{endmembers} latent in the Jasper Ridge\#3:
``\#1 Road'', ``\#2 Soil'', ``\#3 Water'' and ``\#4 Tree'', as shown
in Fig.\,\ref{fig:GT_JasperRidge=0000233}.

\subsection{HYDICE Urban and its two subimages}

HYDICE Urban is one of the most widely used hyperspectral image for
the HU research\ \cite{SJia_07_TGRS_SSCBSS,Jia_09_TGRS_ConstainedNMF,qiangqiangYuan_2012_TGRS_HyperImageDenoising,fyzhu_2014_IJPRS_SSNMF,fyzhu_2014_JSTSP_RRLbS,yingWang_2015_TIP_RobustUnmixing,sigurdsson2017sparse,he2016sparsity,hyunsooKim_2007_Bioinformatics_SNMF,he2017total,tong2017region,Qian_11_TGRS_NMF+l1/2,wu2014sparse}.
It was recorded by the HYDICE (Hyperspectral Digital Image Collection
Experiment) sensor in October 1995, whose scene is an urban area at
Copperas Cove, TX, U.S. There are $307\times307$ pixels in this image;
each pixel corresponds to a $2\times2\thinspace m^{2}$ region in
the scene. The image has 210 spectral bands ranging from $400\thinspace nm$
to $2,500\thinspace nm$, resulting in a very high spectral resolution
of $10\thinspace nm$. After removing the badly degraded bands 1\textendash 4,
76, 87, 101\textendash 111, 136\textendash 153 and 198\textendash 210
(due to dense water vapor and atmospheric effects), we remain 162
bands. There are three versions of ground truths for the HYDICE Urban
hyperspectral image:  
\begin{figure}[t]
\noindent \begin{centering}
\subfloat[Urban\#1 and its six endmembers ``\#1 Asphalt'', ``\#2 Grass'', ``\#3
Tree'', ``\#4 Roof$_{1}$'', ``\#5 Roof2/shado'' and ``6 Soil'' respectively.
\label{fig:GT_Urban=0000231}]{\begin{centering}
\includegraphics[width=1\columnwidth]{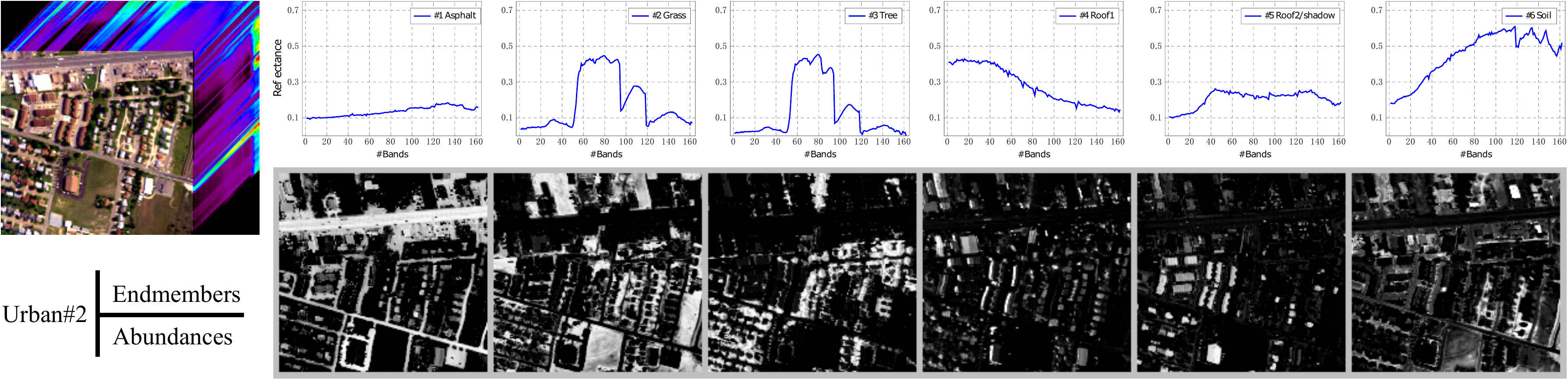}
\par\end{centering}

}
\par\end{centering}

\noindent \begin{centering}
\subfloat[Urban\#2 and its six endmembers ``\#1 Asphalt'', ``\#2 Grass'', ``\#3
Tree'', ``\#4 Roof$_{1}$'', ``\#5 Roof2/shado'' and ``6 Soil'' respectively.
\label{fig:GT_Urban=0000232}]{\begin{centering}
\includegraphics[width=1\columnwidth]{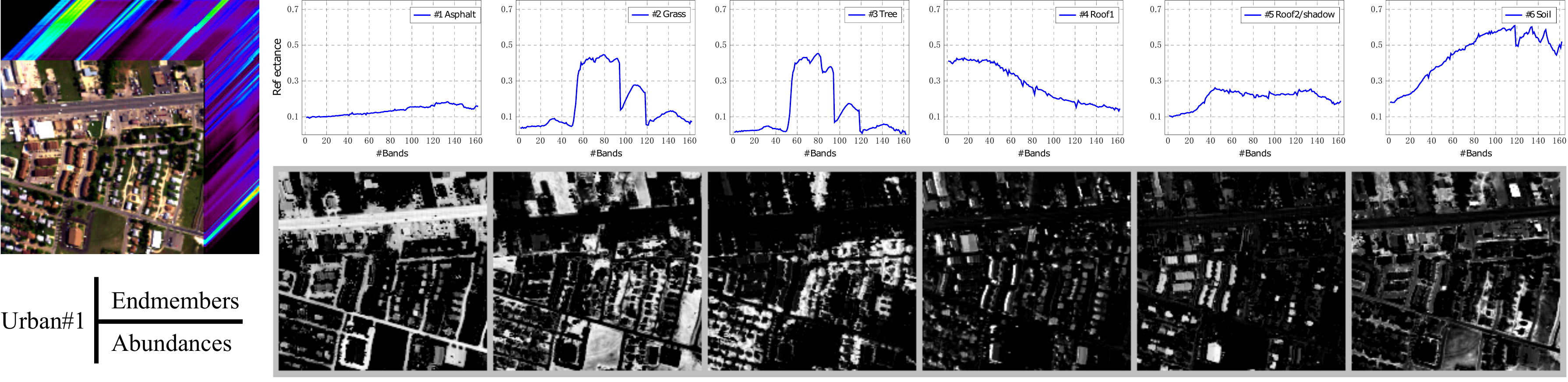}
\par\end{centering}

}
\par\end{centering}

\caption{The two challenging subimages selected from the Urban hyperspectral
image, i.e., Urban\#1 and Urban\#2, respectively. Besides, we provide
the illustration of the \emph{endmembers} and their \emph{abundance}
maps for each subimage\@. }
\end{figure}

\textbf{1)}. In the early HU papers\ \cite{SJia_07_TGRS_SSCBSS,Jia_09_TGRS_ConstainedNMF,fyzhu_2014_IJPRS_SSNMF,fyzhu_2014_JSTSP_RRLbS,yingWang_2015_TIP_RobustUnmixing,Qian_11_TGRS_NMF+l1/2,he2016sparsity,tong2017region},
it is widely accepted that there are four \emph{endmembers} in the
HYDICE Urban. They are ``\#1 Asphalt Road'', ``\#2 Grass'', ``\#3
Tree'' and ``\#4 Roof'', as shown in Fig.\ \ref{fig:GT_Urban}a. 

\textbf{2)}. Recently, the analyse on Urban image become more and
more precise. The ``\#1 Asphalt Road'' is divided into ``\#1 Asphalt''
and ``\#6 Soil'' whereas the ``\#4 Roof'' is divided into ``\#4 Roof$_{1}$''
and ``\#5 Roof$_{2}$/shadow''. In other words, the number of \emph{endmembers}
rises to six\ \cite{Qian_11_TGRS_NMF+l1/2,wu2014sparse}; they are
``\#1 Asphalt'', ``\#2 Grass'', ``\#3 Tree'', ``\#4 Roof$_{1}$'',
``\#5 Roof2/shado'' and ``6 Soil'', as shown in shown in Fig.\ \ref{fig:GT_Urban}c.
 
\begin{figure}[t]
\noindent \begin{centering}
\subfloat[Cuprite]{\noindent \begin{centering}
\includegraphics[height=0.06775\textheight]{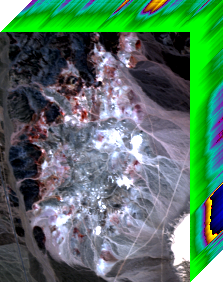}
\par\end{centering}

}\hspace{-0.1cm}\subfloat[The ground truth \emph{endmember} spectrum.]{\begin{centering}
\includegraphics[height=0.06775\textheight]{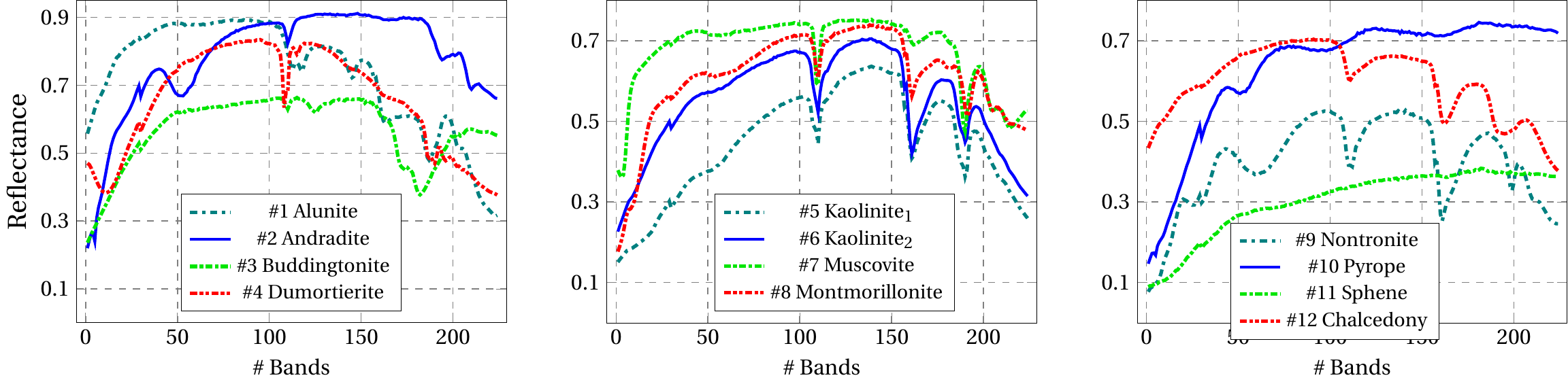}
\par\end{centering}

}
\par\end{centering}

\caption{(a) is the Cuprite hyperspectral image, which is selected in from
the image scene in Fig.\ \ref{fig:7_hyperImages}d, and (b) displays
the ground truth \emph{endmember} spectra. \label{fig:GT_Cuprite}}
\end{figure}

\textbf{3)}. Apart from the above two versions of ground truths, we
introduce a new ground truth that consists of 5 \emph{endmembers}:
``\#1 Asphalt'', ``\#2 Grass'', ``\#3 Tree'', ``\#4 Roof'' and ``5
Soil'', as shown in Fig.\,\ref{fig:GT_Urban}b. Compared with the
first version, the ``\#1 Asphalt Road'' is divided into ``\#1 Asphalt''
and ``\#5 Soil''. Compared with the second version, we merge ``\#4
Roof$_{1}$'' and ``\#5 Roof$_{2}$/shadow'' into ``\#4 Roof''. 

To make it more challenging, we select two subimages from the Urban
image (cf. Fig.\,\ref{fig:7_hyperImages}c). In them, the small objects,
like small houses, vehicles and small grasslands etc., are the main
scene, causing lots of transitional areas. Such case surely leads
to lots of mixed pixels due to the low spatial resolution of hyperspectral
sensors. The first subimage (i.e., Urban\#1) has $160\times168$ pixels,
whose first pixel is the $\left(1,\,1\right)$-th pixel in the original
image. The second subimage (i.e., Urban\#2) starts from the $\left(40,\,1\right)$-th
pixel in the original image. It also has $160\times168$ pixels. There
are six \emph{endmembers}: \textquotedbl{}\#1 Asphalt\textquotedbl{},
\textquotedbl{}\#2 Grass\textquotedbl{}, \textquotedbl{}\#3 Tree\textquotedbl{},
``\#4 Roof$_{1}$'', \textquotedbl{}\#5 Roof2/shadow\textquotedbl{},
and \textquotedbl{}6 Soil\textquotedbl{} in both Urban\#1 (cf., Fig.\,\ref{fig:GT_Urban=0000231})
and Urban\#2 (cf., Fig.\,\ref{fig:GT_Urban=0000232}).

\subsection{Cuprite }

Cuprite is the most benchmark and challenging hyperspectral image
for the HU research\ \cite{agathos2014robust,wei2017unsupervised,bernabe2017parallel,wang2017spatial,wang2016hypergraph,martel2017gpu,fyzhu_2014_TIP_DgS_NMF,XLu_2013_TGRS_ManifoldSparseNMF,LiuXueSong_2011_TGRS_ConstrainedNMF,Qian_11_TGRS_NMF+l1/2,nWang_13_SelectedTopics_EDC-NMF,Jose_05_TGRS_Vca},
which is captured by the AVIRIS sensor. It covers a Cuprite area in
Las Vegas, NV, U.S. There are 224 spectral bands in the Cuprite image,
ranging from $370\thinspace nm$ to $2,480\thinspace nm$. After removing
the noisy bands (i.e., 1\textendash 2 and 221\textendash 224) and
the water absorption bands (i.e., 104\textendash 113 and 148\textendash 167),
it remains 188 bands. In this paper, a subimage of $250\times190$
pixels is considered, which is widely used in the state-of-the-art
HU papers\ \cite{Jose_05_TGRS_Vca,Qian_11_TGRS_NMF+l1/2,XLu_2013_TGRS_ManifoldSparseNMF,fyzhu_2014_TIP_DgS_NMF}.
Please refer to Fig.\,\ref{fig:7_hyperImages}d for the illustration
of its position, image size and scene etc.  
\begin{figure}[t]
\noindent \begin{centering}
\subfloat[Moffett Field\#1 and its three \emph{endmembers}. \label{fig:GT_Moffett Field=0000231}]{\begin{centering}
\includegraphics[width=0.8\columnwidth]{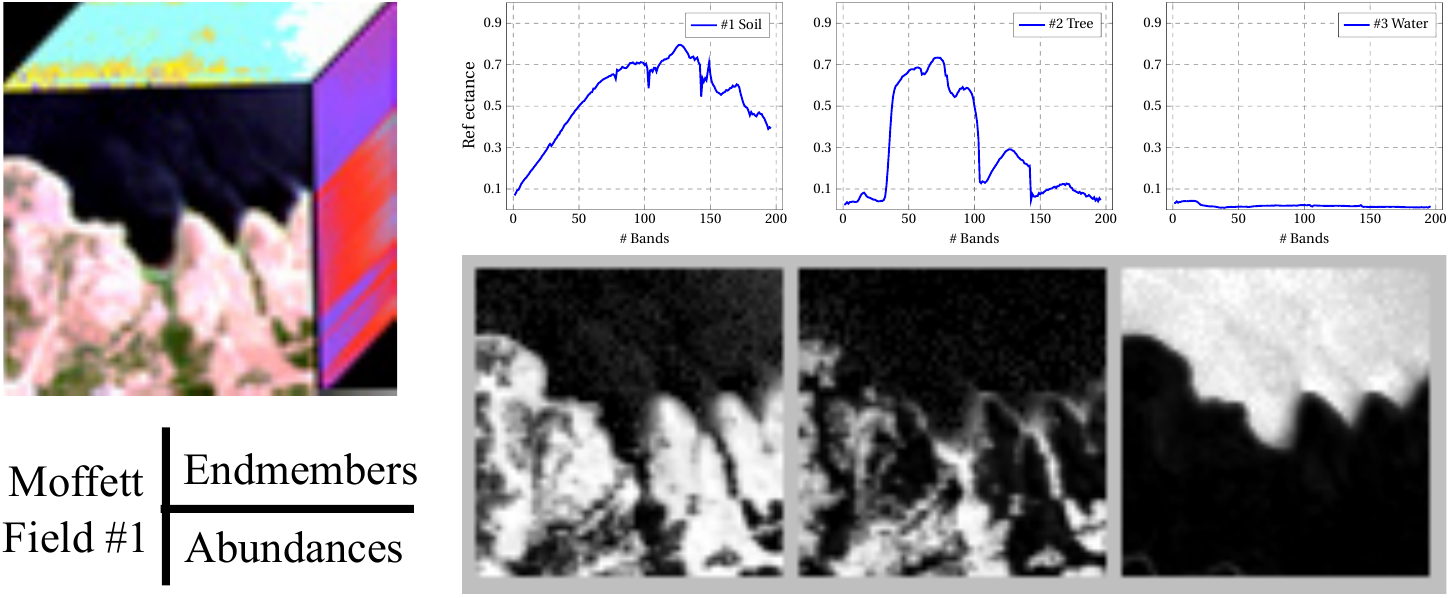}
\par\end{centering}

} 
\par\end{centering}

\noindent \begin{centering}
\vspace{-0.25cm}
\par\end{centering}

\noindent \begin{centering}
\subfloat[Moffett Field\#2 and its three \emph{endmembers}. \label{fig:GT_Moffett Field=0000232}]{\begin{centering}
\includegraphics[width=0.8\columnwidth]{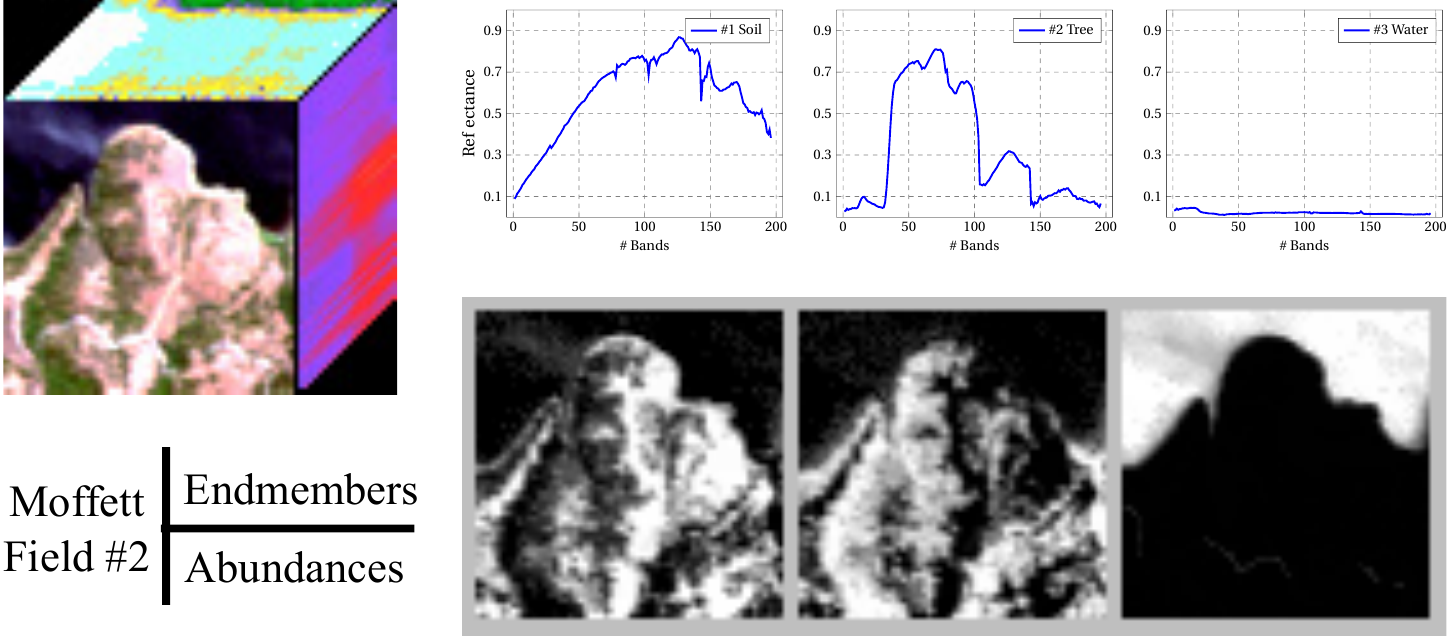}
\par\end{centering}

}
\par\end{centering}

\caption{The two subimages selected from the Moffett Field hyperspectral image.
They are Moffett Field\#1 and Moffett Field\#2, respectively, as shown
in Figs\,\ref{fig:GT_Moffett Field=0000231} and\,\ref{fig:GT_Moffett Field=0000232}.
There are 3 \emph{endmembers}, i.e., ``\#1 Soil'', ``\#2 Tree'' and
``\#3 Water'' in both subimages. Besides, we provide the illustration
of the \emph{endmembers} and their \emph{abundance} maps for each
subimage.\label{fig:GT_2_Moffett Field}}
\end{figure}

In this subimage, there are 14 types of minerals (or \emph{endmembers}),
whose spectra can be obtained from the ENVI software. There are very
minor differences between the variants of the same type of mineral,
for example Kaolinite$_{1}$ and Kaolinite$_{3}$ are very similar
in signature. The researchers have their own thoughts, resulting in
different versions of ground truths. In\ \cite{Jose_05_TGRS_Vca},
there are 14 \emph{endmembers}; while there are 10 \emph{endmembers}
in\ \cite{Qian_11_TGRS_NMF+l1/2}; then Dr. Lu hold that there are
12 \emph{endmembers} in the Cuprite. Here we agree with Dr. Lu's setting.
Please refer to Fig.\,\ref{fig:GT_Cuprite} for the illustration
of the 12 \emph{endmembers}. Because there are small differences in
the setting of \emph{endmembers} among the papers\ \cite{XLu_2013_TGRS_ManifoldSparseNMF,LiuXueSong_2011_TGRS_ConstrainedNMF,Qian_11_TGRS_NMF+l1/2,nWang_13_SelectedTopics_EDC-NMF,Jose_05_TGRS_Vca},
the results of the state-of-the-art methods in those papers might
be different from each other.  
\begin{figure}[t]
\noindent \centering{}\includegraphics[width=0.98\columnwidth]{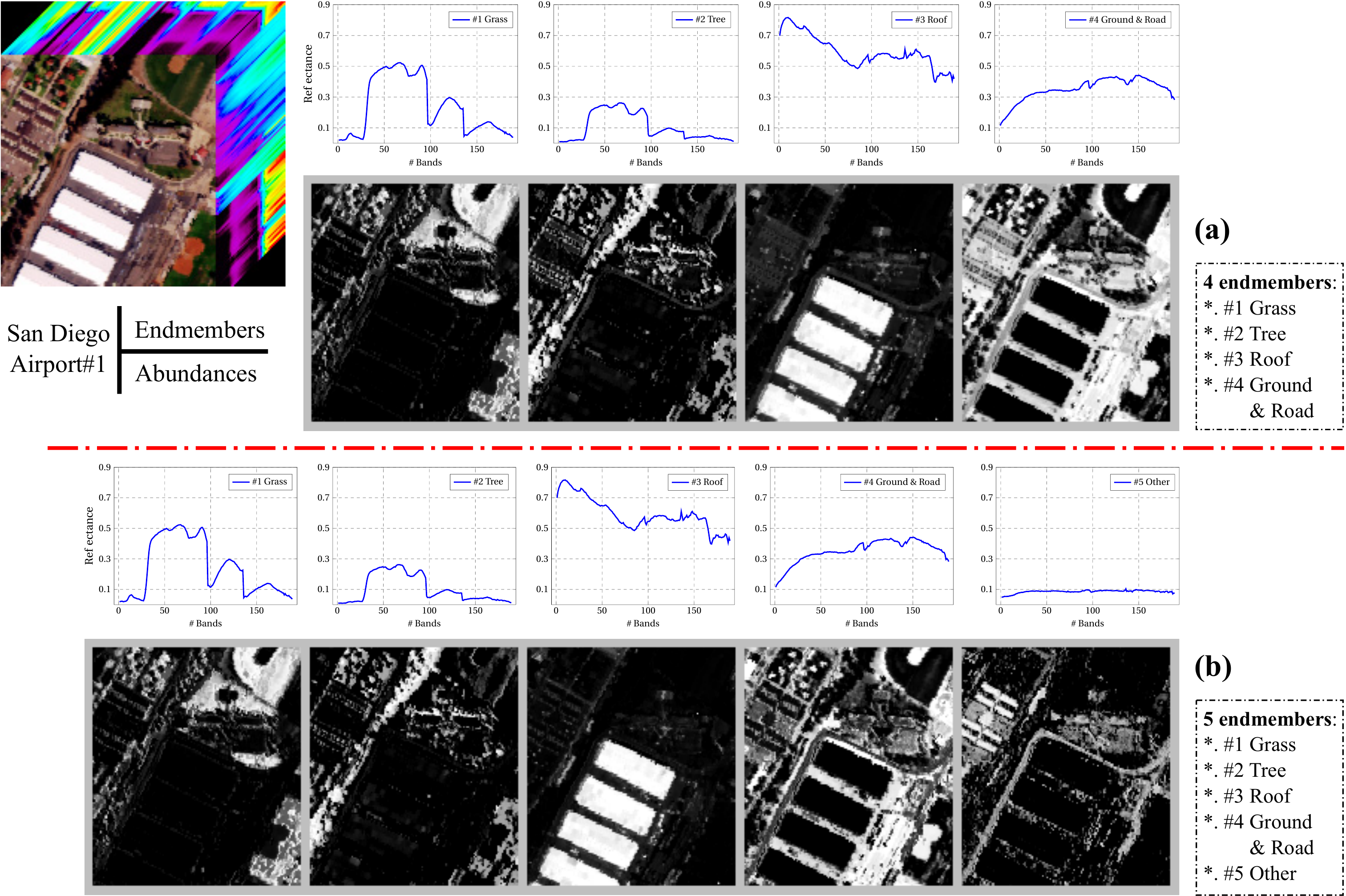}\caption{Two versions of HU ground truths for the San Diego Airport hyperspectral
image: (a) illustrates the ground truth that has 4 \emph{endmembers};
(b) shows the ground truth that has 5 \emph{endmembers}. (best viewed
in color) \label{fig:GT_SanDiegoAirport} }
\end{figure}

\subsection{Moffett Field }

The Moffett Field hyperspectral image is illustrated in Fig.\,\ref{fig:7_hyperImages}e.
It was captured via the AVIRIS sensor by the JPL in 1997 and was used
in a number of state-of-the-art papers\ \cite{li2017transferred,wang2017approximate,zhang2016locally,chen2005archiving,christophe2005quality,dobigeon2007spectral}.
There are $512\times614$ pixels and 224 spectral bands, ranging the
electromagnetic spectra from $400\thinspace nm$ to $2500\thinspace nm$.
The spectral resolution is up to $9.38\thinspace nm$. The full image
can be found in\ \cite{MoffettField_data}. It consists of a large
area of water (that appears in dark pixel at the top of the image)
and a coastal area composed of trees and soil. Due to the water vapor
and atmospheric effects, we remove the noisy spectral bands. After
this process, there remains 196 bands. 

The original image is very large in size and too complex to analyze
because the spatial resolution is low and the object is very small
in the scene. In this paper, we select two subimages for the HU study.
Please refer to Fig.\ref{fig:7_hyperImages}e for their locations,
image sizes and scenes etc. The first subimage (named Moffett Field\#1)
consists of $60\times60$ pixels, whose first (top-left) pixel is
the $\left(54,\,172\right)$-th pixel in the original image. The second
subimage (named Moffett Field\#2) also has $60\times60$ pixels. Its
first pixel is the $\left(62,\,146\right)$-th pixel in the full image.
A similar subimage with Moffett Field\#1 was used in\,\cite{dobigeon2007spectral,eches2011enhancing},
where the authors hold that there are three \emph{endmembers}. In
this paper, we agree with their setting. The three \emph{endmembers}
for both Moffett Field\#1 and Moffett Field\#2 are ``\#1 Soil'', ``\#2
Tree'' and ``\#3 Water'', respectively, as shown in Figs.\,\ref{fig:GT_Moffett Field=0000231}
and\,\ref{fig:GT_Moffett Field=0000232}.

\subsection{San Diego Airport}

San Diego Airport is a popular airborne hyperspectral image collected
by the AVIRIS sensor\ \cite{gillis2013sparse,zhao2014robust,zhao2017progressive,vafadar2017hyperspectral}.
There are $400\times400$ pixels and 224 spectral bands, as illustrated
in Fig.\,\ref{fig:7_hyperImages}f. After removing the noisy bands
with the water vapor and atmospheric effects, including the bands
1\textendash 6, 33\textendash 35, 97, 107\textendash 113, 153\textendash 166,
and 221\textendash 224, there remains 189 spectral bands. In this
paper, we select a subimage for the HU study. It has $160\times140$
pixels. The top-left pixel is the $\left(241,\,261\right)$-th pixel
in the original image, as shown in Fig.\,\ref{fig:7_hyperImages}f.
There are two \emph{endmember} settings: \textbf{(1)} the first setting
has 4 \emph{endmembers}, i.e., ``\#1 Grass'', ``\#2 Tree'', ``\#3
Roof'' and ``\#4 Ground \& Road'', as shown in Fig.\,\ref{fig:GT_SanDiegoAirport}a;
\textbf{(2)} the second setting has 5 \emph{endmembers}, i.e., ``\#1
Grass'', ``\#2 Tree'', ``\#3 Roof'', ``\#4 Ground \& Road'' and ``\#5
Other'', as shown in Fig.\,\ref{fig:GT_SanDiegoAirport}b.  
\begin{figure}[t]
\noindent \begin{centering}
\subfloat[WDM\#1 and its six \emph{endmembers}. \label{fig:Washington-DC-Mall=0000231}]{\begin{centering}
\includegraphics[width=0.98\columnwidth]{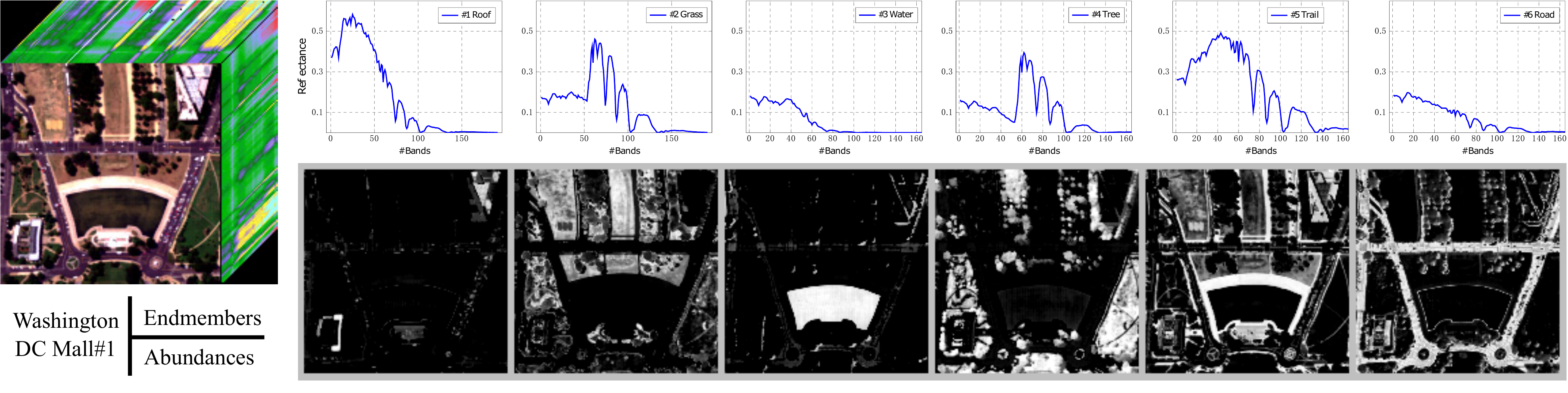}
\par\end{centering}

}
\par\end{centering}

\noindent \begin{centering}
\subfloat[WDM\#2 and its six \emph{endmembers}. \label{fig:Washington-DC-Mall=0000232}]{\begin{centering}
\includegraphics[width=0.98\columnwidth]{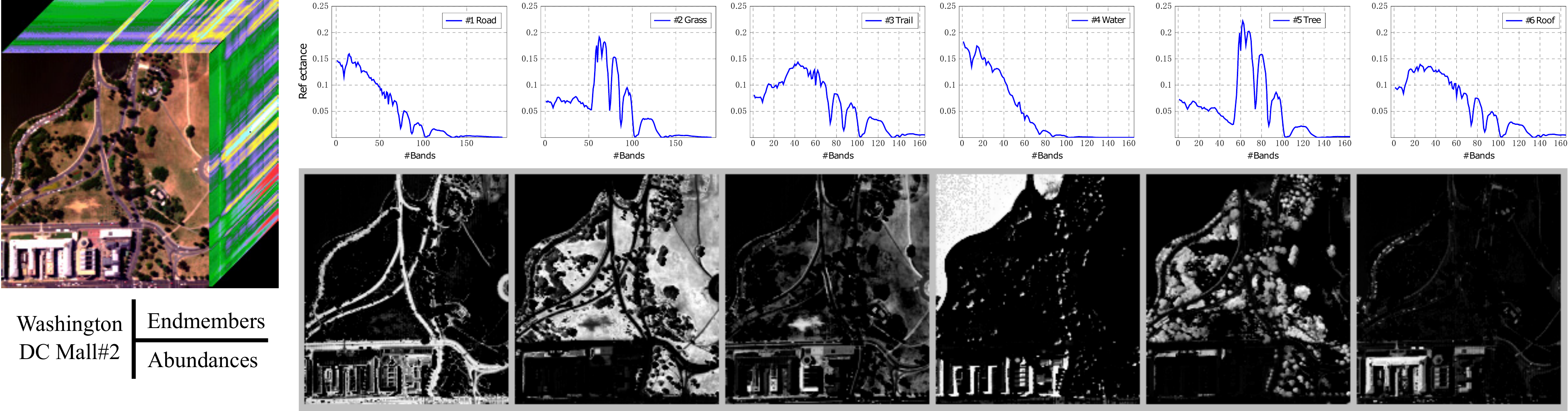}
\par\end{centering}

}
\par\end{centering}

\caption{Two challenging subimages selected from the Washington DC Mall image:
WDM\#1 and WDM\#2 as shown in Figs\ \ref{fig:Washington-DC-Mall=0000231}
and\,\ref{fig:Washington-DC-Mall=0000232}. For each subimage, we
give the illustration of the ground truth \emph{endmembers} and \emph{abundances}\@.
\label{fig:Washington-DC-Mall}}
\end{figure}

\subsection{Washington DC Mall (WDM)}

The \textbf{W}ashington \textbf{D}C \textbf{M}all (i.e., WDM), an
airborne hyperspectral image flightline over the Washington DC Mall
(cf. Fig.\,\ref{fig:Washington-DC-Mall}g), is used to verify a lot
of research works\ \cite{xu2017using,tong2017region,wang2016hypergraph,Jia_09_TGRS_ConstainedNMF,david2002hyperspectral,nWang_13_SelectedTopics_EDC-NMF}.
It was collected by the HYDICE sensor. The full image can be downloaded
from the webpage\footnote{\href{https://engineering.purdue.edu/~biehl/MultiSpec/hyperspectral.html}{https://engineering.purdue.edu/$\sim$biehl/MultiSpec/hyperspectral.html}}.
There are $1208\times307$ pixels and $210$ spectral channels, covering
the the electromagnetic spectra from $400\thinspace nm$ to $2400\thinspace nm$.
The spectral resolution is up to $9.52\thinspace nm$. After removing
the bands with water vapor and atmospheric effects, including 103\textendash 106,
138\textendash 148, and 207\textendash 210, there remains 191 bands\,\cite{wang2016hypergraph}. 

In this paper, we select two subimages for the HU study, i.e., WDM\#1
and WDM\#2 respectively. Please refer to Fig.\,\ref{fig:Washington-DC-Mall}g
and Table\ \ref{tab:18_realHyperDatasets} for their information
like locations and scenes etc. The first subimage (i.e. WDM\#1) has
$150\times150$ pixels, whose first pixel starts from the $\left(549,\,160\right)$-th
pixel in the original image. WDM\#1 has been used in\ \cite{nWang_13_SelectedTopics_EDC-NMF},
where they provide the rough locations of the six \emph{endmember};
however, they did not share the ground truth on the Internet. The
six \emph{endmembers }are ``\#1 Roof'', ``\#2 Grass'', ``\#3 Water'',
``\#4 Tree'', ``\#5 Trail'' and ``\#6 Road'' respectively, as shown
in Fig.\,\ref{fig:Washington-DC-Mall=0000231}. The second subimage
(i.e. WDM\#2) has $180\times160$ pixels, whose left-top pixel is
the $\left(945,\,90\right)$-th pixel in the original image. The six
\emph{endmembers} are ``\#1 Road'', ``\#2 Grass'', ``\#3 Trail'',
``\#4 Water'', ``\#5 Tree'' and ``\#6 Roof'' respectively, as shown
in Fig.\,\ref{fig:Washington-DC-Mall=0000232}.

\section{The Simulated Hyperspectral Image\label{sec:simulated_hyperspectral_images}}

\begin{figure}[t]
\noindent \begin{centering}
\includegraphics[width=0.99\columnwidth]{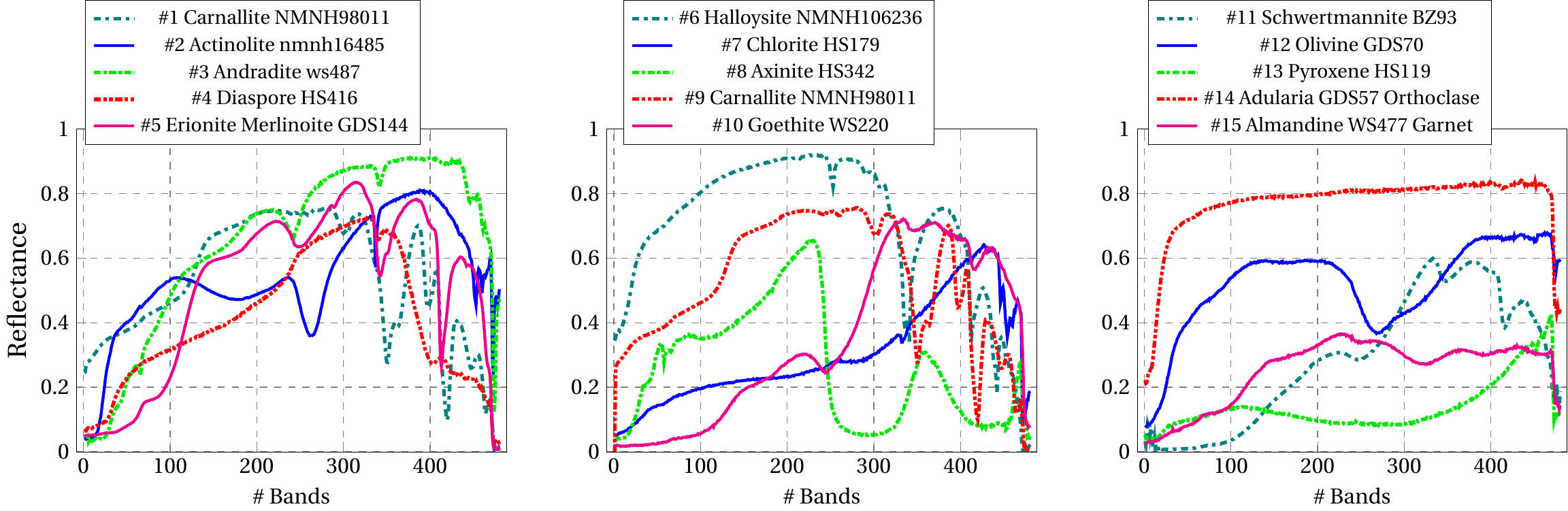}
\par\end{centering}

\caption{The signature of 15 candidate \emph{endmembers} used for the simulation
of hyperspectral images. Each signature is observed at 480 spectral
bands. \label{fig:synthetic_15_signatures}}
\end{figure}

This section provides the method to generate a complex synthetic image\ \cite{Miao_07_ITGRS_NMFMVC,Jia_09_TGRS_ConstainedNMF,yingWang_2015_TIP_RobustUnmixing,Qian_11_TGRS_NMF+l1/2,wang2016parallel,tong2016nonnegative,tong2017region,qian2017matrix}
for the HU research. It consists of two steps. First, 15 true spectra
are chosen from the United States Geological Survey (USGS) digital
spectral library as the candidate \emph{endmembers}. Their signature
are shown in Fig.\,\ref{fig:synthetic_15_signatures}. The first
$K$ spectra are selected as the true \emph{endmember}, where $2\leq K\leq15$
is the number of \emph{endmembers.} 

Second, we use the method in\ \cite{Miao_07_ITGRS_NMFMVC,Jia_09_TGRS_ConstainedNMF,yingWang_2015_TIP_RobustUnmixing,Qian_11_TGRS_NMF+l1/2,tong2016nonnegative,qian2017matrix}
to generate the \emph{abundance} maps, which has the following steps:

\textbf{1)}. We divide an image of $z^{2}\times z^{2}\mbox{\ensuremath{\left(z\in\mtbbZ^{+}\right)}}$
pixels into $z\times z$ regions. The default setting is $z=8$. 

\textbf{2)}. Each region is filled up with a type of ground cover,
which means the \emph{abundance} of that \emph{endmember} is 1 in
the whole region. Note that, this process is randomly implemented.

\textbf{3)}. To generate the mixed pixels, we use the $\left(z+1\right)\times\left(z+1\right)$
spatial low-pass filter. Such process causes large transitional areas,
where there are lots of mixed pixels.

\textbf{4)}. To remove pure pixels and generate sparse \emph{abundances},
we replace the pixels whose largest \emph{abundance} is larger than
0.8 with a mixture of two \emph{endmembers} with equal \emph{abundances}.

Now both \emph{endmembers $\mhbfM$} and \emph{abundances} $\mhbfA$
have been generated. Based on them, a perfect hyperspectral image
is created as $\mhbfY=\mhbfM\mhbfA$. To simulate a more real remote
sensing image that contains complex noises, we add the zero-mean Gaussian
(or some other) noise to the generate the image data. The Signal-to-Noise
Ratio (SNR) is defined as follows:
\begin{equation}
\text{SNR}=10\log_{10}\left\{ \frac{\mtexpect\left[\mtbfy^{\top}\mtbfy\right]}{\mtexpect\left[\mtbfn^{\top}\mtbfn\right]}\right\} .
\end{equation}

With the above synthetic hyperspectral image, we can verify the HU
algorithmss from the following four perspectives:
\begin{itemize}
\item Robustness analysis to the noise corruptions, e.g., $\text{SNR\ensuremath{\in\left\{  \infty,35,\cdots,10\thinspace\text{dB}\right\} } }.$
\item Generalization to the pixel number, which is set by $z$. 
\item Robustness to the \emph{endmember} number\emph{ $\left\{ K\!\mid\!2\!\leq\!K\!\leq\!15\right\} $. }
\item Generalization to the number of bands in the image.
\end{itemize}
We will provide the full codes for this simulated dataset.

\begin{table*}[t]
\begin{centering}
\caption{The performances of five benchmark HU algorithms on the Samson\#1,
Samson\#2 and Samson\#3 hyperspectral images. \emph{\label{tab:rst1_Samson_1=0000262=0000263}}}
\vspace{-0.25cm}
 \resizebox{1.75\columnwidth}{!}{%
\begin{tabular}{|l||l|ccccc|ccccc|}
\hline 
\multirow{6}{*}{\begin{turn}{90}
Samson \#1
\end{turn}} &
\multirow{2}{*}{\emph{Endmembers}} &
\multicolumn{5}{c|}{\textbf{S}pectral \textbf{A}ngle \textbf{D}istance\textbf{ }(\textbf{SAD})} &
\multicolumn{5}{c|}{\textbf{R}oot \textbf{M}ean \textbf{S}quare \textbf{E}rror (\textbf{RMSE})}\tabularnewline
 &  & VCA &
NMF &
$\ell_{1}$-NMF &
$\ell_{1/2}$-NMF &
G-NMF &
VCA &
NMF &
$\ell_{1}$-NMF &
$\ell_{1/2}$-NMF &
G-NMF\tabularnewline
\cline{2-12} 
 & \#1 Soil &
0.1531  &
0.0861  &
0.0582  &
0.0902  &
0.0529  &
0.1765  &
0.1110  &
0.0930  &
0.0990  &
0.0993 \tabularnewline
 & \#2 Tree &
0.0487  &
0.0556  &
0.0545  &
0.0410  &
0.0576  &
0.1372  &
0.0886  &
0.0798  &
0.0919  &
0.0734 \tabularnewline
 & \#3 Water &
0.1296  &
0.1560  &
0.1233  &
0.0410  &
0.1609  &
0.1748  &
0.0757  &
0.0511  &
0.0344  &
0.0750 \tabularnewline
 & Avg. &
0.1105  &
0.0992  &
0.0787  &
0.0574  &
0.0905  &
0.1628  &
0.0918  &
0.0746  &
0.0751  &
0.0826 \tabularnewline
\hline 
\hline 
\multirow{6}{*}{\begin{turn}{90}
Samson \#2
\end{turn}} &
\multirow{2}{*}{\emph{Endmembers}} &
\multicolumn{5}{c|}{\textbf{S}pectral \textbf{A}ngle \textbf{D}istance\textbf{ }(\textbf{SAD})} &
\multicolumn{5}{c|}{\textbf{R}oot \textbf{M}ean \textbf{S}quare \textbf{E}rror (\textbf{RMSE})}\tabularnewline
 &  & VCA &
NMF &
$\ell_{1}$-NMF &
$\ell_{1/2}$-NMF &
G-NMF &
VCA &
NMF &
$\ell_{1}$-NMF &
$\ell_{1/2}$-NMF &
G-NMF\tabularnewline
\cline{2-12} 
 & \#1 Soil &
0.4239  &
0.2793  &
0.1780  &
0.2074  &
0.2647  &
0.1504  &
0.1633  &
0.1425  &
0.1719  &
0.1583 \tabularnewline
 & \#2 Tree &
0.1118  &
0.1150  &
0.0542  &
0.0559  &
0.1161  &
0.1483  &
0.1710  &
0.1341  &
0.1683  &
0.1663 \tabularnewline
 & \#3 Water &
0.0662  &
0.0804  &
0.0778  &
0.0731  &
0.0821  &
0.1055  &
0.0610  &
0.0360  &
0.0395  &
0.0610 \tabularnewline
 & Avg. &
0.2006  &
0.1582  &
0.1033  &
0.1121  &
0.1543  &
0.1347  &
0.1318  &
0.1042  &
0.1266  &
0.1285 \tabularnewline
\hline 
\hline 
\multirow{6}{*}{\begin{turn}{90}
Samson \#3
\end{turn}} &
\multirow{2}{*}{\emph{Endmembers}} &
\multicolumn{5}{c|}{\textbf{S}pectral \textbf{A}ngle \textbf{D}istance\textbf{ }(\textbf{SAD})} &
\multicolumn{5}{c|}{\textbf{R}oot \textbf{M}ean \textbf{S}quare \textbf{E}rror (\textbf{RMSE})}\tabularnewline
 &  & VCA &
NMF &
$\ell_{1}$-NMF &
$\ell_{1/2}$-NMF &
G-NMF &
VCA &
NMF &
$\ell_{1}$-NMF &
$\ell_{1/2}$-NMF &
G-NMF\tabularnewline
\cline{2-12} 
 & \#1 Soil &
0.0454  &
0.0422  &
0.0315  &
0.0316  &
0.0509  &
0.0801  &
0.0792  &
0.0326  &
0.0498  &
0.0870 \tabularnewline
 & \#2 Tree &
0.0600  &
0.0714  &
0.0425  &
0.0429  &
0.0781  &
0.0771  &
0.0802  &
0.0319  &
0.0427  &
0.0895 \tabularnewline
 & \#3 Water &
0.1134  &
0.1155  &
0.1129  &
0.1135  &
0.1158  &
0.0217  &
0.0263  &
0.0072  &
0.0134  &
0.0321 \tabularnewline
 & Avg. &
0.0729  &
0.0764  &
0.0623  &
0.0627  &
0.0816  &
0.0596  &
0.0619  &
0.0239  &
0.0353  &
0.0695 \tabularnewline
\hline 
\end{tabular}}
\par\end{centering}

\centering{}\vspace{-0.25cm}
\end{table*}

\begin{table}[t]
\begin{centering}
\caption{The SAD performances of five benchmark HU algorithms on the Cuprite
hyperspectral images. \label{tab:rst2_Cuprite}}
\vspace{-0.25cm}
 \resizebox{0.98\columnwidth}{!}{%
\begin{tabular}{|l|ccccc|}
\hline 
\multirow{2}{*}{\emph{Endmembers}} &
\multicolumn{5}{c|}{\textbf{S}pectral \textbf{A}ngle \textbf{D}istance (\textbf{SAD})}\tabularnewline
 & VCA &
NMF &
$\ell_{1}$-NMF &
$\ell_{1/2}$-NMF &
G-NMF\tabularnewline
\hline 
\#1 Alunite &
0.1584  &
0.1640  &
0.1618  &
0.1458  &
0.1632 \tabularnewline
\#2 Andradite &
0.0773  &
0.1166  &
0.1253  &
0.0783  &
0.1279 \tabularnewline
\#3 Buddingtonite &
0.0999  &
0.1343  &
0.1391  &
0.1456  &
0.1406 \tabularnewline
\#4 Dumortierite &
0.1033  &
0.1202  &
0.1212  &
0.1067  &
0.1211 \tabularnewline
\#5 Kaolinite$_{1}$ &
0.0892  &
0.1042  &
0.0951  &
0.1004  &
0.1001 \tabularnewline
\#6 Kaolinite$_{2}$ &
0.0682  &
0.1120  &
0.1024  &
0.0763  &
0.1125 \tabularnewline
\#7 Muscovite &
0.1912  &
0.2440  &
0.2490  &
0.2228  &
0.2489 \tabularnewline
\multicolumn{1}{|l|}{\#8 Montmorillonite} &
0.0721  &
0.1165  &
0.0980  &
0.0769  &
0.1209 \tabularnewline
\#9 Nontronite &
0.0784  &
0.1115  &
0.1036  &
0.0940  &
0.1232 \tabularnewline
\#10 Pyrope &
0.0907  &
0.0966  &
0.0935  &
0.0831  &
0.1045 \tabularnewline
\#11 Sphene &
0.0702  &
0.0952  &
0.0903  &
0.0788  &
0.0923 \tabularnewline
\#12 Chalcedony &
0.0870  &
0.1811  &
0.1719  &
0.1542  &
0.1513 \tabularnewline
Avg. &
0.0988  &
0.1330  &
0.1293  &
0.1136  &
0.1339 \tabularnewline
\hline 
\end{tabular}}
\par\end{centering}

\vspace{-0.25cm}
\end{table}

\begin{table*}[t]
\begin{centering}
\caption{The performances of five benchmark HU algorithms on the HYDICE Urban
image, which has three versions of ground truths.\label{tab:rst3_Urban_end4,5,6}}
\vspace{-0.25cm}
\resizebox{1.75\columnwidth}{!}{%
\begin{tabular}{|l||l|ccccc|ccccc|}
\hline 
\multirow{7}{*}{\begin{turn}{90}
V1: 4 \emph{endmembers}
\end{turn}} &
\multirow{2}{*}{\emph{Endmembers}} &
\multicolumn{5}{c|}{\textbf{S}pectral \textbf{A}ngle \textbf{D}istance\textbf{ }(\textbf{SAD})} &
\multicolumn{5}{c|}{\textbf{R}oot \textbf{M}ean \textbf{S}quare \textbf{E}rror (\textbf{RMSE})}\tabularnewline
 &  & VCA &
NMF &
$\ell_{1}$-NMF &
$\ell_{1/2}$-NMF &
G-NMF &
VCA &
NMF &
$\ell_{1}$-NMF &
$\ell_{1/2}$-NMF &
G-NMF\tabularnewline
\cline{2-12} 
 & \#1 Asphalt &
0.2152  &
0.2114  &
0.1548  &
0.1349  &
0.2086  &
0.2884  &
0.2041  &
0.2279  &
0.3225  &
0.2056 \tabularnewline
 & \#2 Grass &
0.3547  &
0.3654  &
0.2876  &
0.0990  &
0.3716  &
0.3830  &
0.2065  &
0.2248  &
0.3387  &
0.2078 \tabularnewline
 & \#3 Tree &
0.2115  &
0.1928  &
0.0911  &
0.0969  &
0.1934  &
0.2856  &
0.1870  &
0.1736  &
0.2588  &
0.1824 \tabularnewline
 & \#4 Roof &
0.7697  &
0.7370  &
0.7335  &
0.5768  &
0.7377  &
0.1638  &
0.1395  &
0.1861  &
0.1782  &
0.1389 \tabularnewline
 & Avg. &
0.3877  &
0.3767  &
0.3168  &
0.2269  &
0.3778  &
0.2802  &
0.1843  &
0.2031  &
0.2746  &
0.1837 \tabularnewline
\hline 
\hline 
\multirow{8}{*}{\begin{turn}{90}
V2: 5 \emph{endmembers}
\end{turn}} &
\multirow{2}{*}{\emph{Endmembers}} &
\multicolumn{5}{c|}{\textbf{S}pectral \textbf{A}ngle \textbf{D}istance\textbf{ }(\textbf{SAD})} &
\multicolumn{5}{c|}{\textbf{R}oot \textbf{M}ean \textbf{S}quare \textbf{E}rror (\textbf{RMSE})}\tabularnewline
 &  & VCA &
NMF &
$\ell_{1}$-NMF &
$\ell_{1/2}$-NMF &
G-NMF &
VCA &
NMF &
$\ell_{1}$-NMF &
$\ell_{1/2}$-NMF &
G-NMF\tabularnewline
\cline{2-12} 
 & \#1 Asphalt &
0.3009  &
0.3284  &
0.1664  &
0.1837  &
0.3502  &
0.2898 &
0.2941 &
0.3080  &
0.3408  &
0.2944 \tabularnewline
 & \#2 Grass &
0.3547  &
0.4575  &
0.1826  &
0.1143  &
0.5539  &
0.3634 &
0.3417 &
0.2048  &
0.3099  &
0.3823 \tabularnewline
 & \#3 Tree &
0.4694  &
0.2337  &
0.1335  &
0.0658  &
0.2402  &
0.3092 &
0.2513 &
0.1245  &
0.2944  &
0.2887 \tabularnewline
 & \#4 Roof &
0.4066  &
0.4134  &
0.4801  &
0.0834  &
0.4340  &
0.2202 &
0.1548 &
0.1101  &
0.1751  &
0.1581 \tabularnewline
 & \#5 Soil &
1.0638  &
1.0478  &
1.0948  &
0.0290  &
0.9651  &
0.2791 &
0.2755 &
0.2797  &
0.3125  &
0.2796 \tabularnewline
 & Avg. &
0.5191  &
0.4962  &
0.4115  &
0.0952  &
0.5087  &
0.2923 &
0.2635 &
0.2054  &
0.2865  &
0.2806 \tabularnewline
\hline 
\hline 
\multirow{10}{*}{\begin{turn}{90}
V3: 6 \emph{endmembers}
\end{turn}} &
\multirow{2}{*}{\emph{Endmembers}} &
\multicolumn{5}{c|}{\textbf{S}pectral \textbf{A}ngle \textbf{D}istance\textbf{ }(\textbf{SAD})} &
\multicolumn{5}{c|}{\textbf{R}oot \textbf{M}ean \textbf{S}quare \textbf{E}rror (\textbf{RMSE})}\tabularnewline
 &  & VCA &
NMF &
$\ell_{1}$-NMF &
$\ell_{1/2}$-NMF &
G-NMF &
VCA &
NMF &
$\ell_{1}$-NMF &
$\ell_{1/2}$-NMF &
G-NMF\tabularnewline
\cline{2-12} 
 & \#1 Asphalt &
0.2441  &
0.3322  &
0.2669  &
0.3092  &
0.3274  &
0.2555  &
0.2826  &
0.2803  &
0.3389  &
0.2787 \tabularnewline
 & \#2 Grass &
0.3058  &
0.4059  &
0.3294  &
0.0792  &
0.3787  &
0.2696  &
0.3536  &
0.2937  &
0.2774  &
0.3184 \tabularnewline
 & \#3 Tree &
0.6371  &
0.2558  &
0.2070  &
0.0623  &
0.2629  &
0.3212  &
0.2633  &
0.1859  &
0.2701  &
0.2404 \tabularnewline
 & \#4 Roof1 &
0.2521  &
0.3701  &
0.4370  &
0.0680  &
0.3264  &
0.2500  &
0.1662  &
0.1358  &
0.1541  &
0.1725 \tabularnewline
 & \#5 Roof2/ &
\multirow{2}{*}{0.7451 } &
\multirow{2}{*}{0.6223 } &
\multirow{2}{*}{0.5330 } &
\multirow{2}{*}{0.1870 } &
\multirow{2}{*}{0.6831 } &
\multirow{2}{*}{0.2157 } &
\multirow{2}{*}{0.1603 } &
\multirow{2}{*}{0.2136 } &
\multirow{2}{*}{0.1660 } &
\multirow{2}{*}{0.1441 }\tabularnewline
 & \ \ \ Shadow &  &  &  &  &  &  &  &  &  & \tabularnewline
 & \#6 Soil &
1.1061  &
0.9978  &
1.0371  &
0.0287  &
0.9859  &
0.2714  &
0.2505  &
0.2594  &
0.3542  &
0.2547 \tabularnewline
 & Avg. &
0.5484  &
0.4974  &
0.4684  &
0.1224  &
0.4941  &
0.2639  &
0.2461  &
0.2281  &
0.2601  &
0.2348 \tabularnewline
\hline 
\end{tabular}}
\par\end{centering}

\begin{centering}
\vspace{0.25cm}

\par\end{centering}

\begin{centering}
\caption{The performances of five benchmark HU algorithms on the Urban\#1 and
Urbann\#2 hyperspectral sub-images. \label{tab:rst4_Urban_R1=000026R2}}
\vspace{-0.25cm}
 \resizebox{1.75\columnwidth}{!}{%
\begin{tabular}{|l||l|ccccc|ccccc|}
\hline 
\multirow{10}{*}{\begin{turn}{90}
Urban \#1
\end{turn}} &
\multirow{2}{*}{\emph{Endmembers}} &
\multicolumn{5}{c|}{\textbf{S}pectral \textbf{A}ngle \textbf{D}istance\textbf{ }(\textbf{SAD})} &
\multicolumn{5}{c|}{\textbf{R}oot \textbf{M}ean \textbf{S}quare \textbf{E}rror (\textbf{RMSE})}\tabularnewline
 &  & VCA &
NMF &
$\ell_{1}$-NMF &
$\ell_{1/2}$-NMF &
G-NMF &
VCA &
NMF &
$\ell_{1}$-NMF &
$\ell_{1/2}$-NMF &
G-NMF\tabularnewline
\cline{2-12} 
 & \#1 Asphalt &
0.2166  &
0.3172  &
0.1843  &
0.2761  &
0.3273  &
0.2962  &
0.2544  &
0.2663  &
0.2475  &
0.2488 \tabularnewline
 & \#2 Grass &
1.3196  &
0.5520  &
1.1585  &
0.6055  &
0.4975  &
0.4199  &
0.2727  &
0.3864  &
0.2982  &
0.2720 \tabularnewline
 & \#3 Tree &
0.1696  &
0.1221  &
0.0708  &
0.1034  &
0.1188  &
0.1934  &
0.1648  &
0.2271  &
0.1602  &
0.1709 \tabularnewline
 & \#4 Roof1 &
0.2475  &
0.3403  &
0.3632  &
0.3537  &
0.3686  &
0.3176  &
0.1871  &
0.1411  &
0.1782  &
0.1695 \tabularnewline
 & \#5 Roof2/ &
\multirow{2}{*}{0.9453 } &
\multirow{2}{*}{0.6446 } &
\multirow{2}{*}{0.3751 } &
\multirow{2}{*}{0.6482 } &
\multirow{2}{*}{0.6075 } &
\multirow{2}{*}{0.1965 } &
\multirow{2}{*}{0.1957 } &
\multirow{2}{*}{0.2693 } &
\multirow{2}{*}{0.2080 } &
\multirow{2}{*}{0.2020 }\tabularnewline
 & \ \ \ Shadow &  &  &  &  &  &  &  &  &  & \tabularnewline
 & \#6 Soil &
0.3329  &
0.4991  &
0.2179  &
0.4191  &
0.4560  &
0.2409  &
0.2661  &
0.2509  &
0.2843  &
0.2696 \tabularnewline
 & Avg. &
0.5386  &
0.4125  &
0.3950  &
0.4010  &
0.3959  &
0.2774  &
0.2235  &
0.2568  &
0.2294  &
0.2222 \tabularnewline
\hline 
\multirow{10}{*}{\begin{turn}{90}
Urban \#2
\end{turn}} &
\multirow{2}{*}{\emph{Endmembers}} &
\multicolumn{5}{c|}{\textbf{S}pectral \textbf{A}ngle \textbf{D}istance\textbf{ }(\textbf{SAD})} &
\multicolumn{5}{c|}{\textbf{R}oot \textbf{M}ean \textbf{S}quare \textbf{E}rror (\textbf{RMSE})}\tabularnewline
 &  & VCA &
NMF &
$\ell_{1}$-NMF &
$\ell_{1/2}$-NMF &
G-NMF &
VCA &
NMF &
$\ell_{1}$-NMF &
$\ell_{1/2}$-NMF &
G-NMF\tabularnewline
\cline{2-12} 
 & \#1 Asphalt &
0.2112  &
0.2693  &
0.2319  &
0.4158  &
0.2874  &
0.2724  &
0.2710  &
0.2617  &
0.4366  &
0.2488 \tabularnewline
 & \#2 Grass &
1.2238  &
0.2990  &
0.3400  &
0.2348  &
0.3526  &
0.3673  &
0.2341  &
0.2662  &
0.2809  &
0.2430 \tabularnewline
 & \#3 Tree &
0.1778  &
0.1263  &
0.1203  &
0.1037  &
0.1212  &
0.1910  &
0.1636  &
0.1776  &
0.2120  &
0.1732 \tabularnewline
 & \#4 Roof1 &
0.2012  &
0.4419  &
0.4462  &
0.4320  &
0.4304  &
0.3693  &
0.1360  &
0.1239  &
0.1341  &
0.1359 \tabularnewline
 & \#5 Roof2/ &
\multirow{2}{*}{1.0188 } &
\multirow{2}{*}{0.5260 } &
\multirow{2}{*}{0.5151 } &
\multirow{2}{*}{0.3300 } &
\multirow{2}{*}{0.5371 } &
\multirow{2}{*}{0.1857 } &
\multirow{2}{*}{0.2244 } &
\multirow{2}{*}{0.2249 } &
\multirow{2}{*}{0.2448 } &
\multirow{2}{*}{0.2030 }\tabularnewline
 & \ \ \ Shadow &  &  &  &  &  &  &  &  &  & \tabularnewline
 & \#6 Soil &
0.4134  &
0.5414  &
0.4434  &
0.0786  &
0.5253  &
0.2235  &
0.2619  &
0.2675  &
0.4651  &
0.2535 \tabularnewline
 & Avg. &
0.5410  &
0.3673  &
0.3495  &
0.2658  &
0.3757  &
0.2682  &
0.2152  &
0.2203  &
0.2956  &
0.2096 \tabularnewline
\hline 
\end{tabular}}\vspace{-0.25cm}

\par\end{centering}

\centering{}
\end{table*}

\begin{table*}[t]
\begin{centering}
\caption{The HU results of five benchmark HU algorithms on the JasperRidge\#1,
JasperRidge\#2 and JasperRidge\#3 hyperspectral images.\label{tab:rst5_JasperRidge_1=0000262=0000263}}
\vspace{-0.25cm}
\resizebox{1.75\columnwidth}{!}{%
\begin{tabular}{|l||l|ccccc|ccccc|}
\hline 
\multirow{8}{*}{\begin{turn}{90}
Jasper Ridge \#1
\end{turn}} &
\multirow{2}{*}{\emph{Endmembers}} &
\multicolumn{5}{c|}{\textbf{S}pectral \textbf{A}ngle \textbf{D}istance\textbf{ }(\textbf{SAD})} &
\multicolumn{5}{c|}{\textbf{R}oot \textbf{M}ean \textbf{S}quare \textbf{E}rror (\textbf{RMSE})}\tabularnewline
 &  & VCA &
NMF &
$\ell_{1}$-NMF &
$\ell_{1/2}$-NMF &
G-NMF &
VCA &
NMF &
$\ell_{1}$-NMF &
$\ell_{1/2}$-NMF &
G-NMF\tabularnewline
\cline{2-12} 
 & \#1 Tree &
0.3381  &
0.1325  &
0.0553  &
0.0458  &
0.0743  &
0.2651  &
0.1327  &
0.0928  &
0.1032  &
0.0752 \tabularnewline
 & \#2 Water &
0.2573  &
0.2823  &
0.0300  &
0.0303  &
0.2923  &
0.2711  &
0.1384  &
0.0714  &
0.0911  &
0.1366 \tabularnewline
 & \#3 Soil &
0.3062  &
0.2125  &
0.1318  &
0.0534  &
0.2005  &
0.2567  &
0.2221  &
0.2536  &
0.2631  &
0.1837 \tabularnewline
 & \#4 Road &
0.6053  &
0.4843  &
0.6352  &
0.5525  &
0.5411  &
0.2361  &
0.1876  &
0.2064  &
0.2389  &
0.1908 \tabularnewline
 & \#5 Other &
0.7156  &
0.6463  &
1.0503  &
0.8187  &
0.6742  &
0.2842  &
0.1601  &
0.1357  &
0.1451  &
0.1669 \tabularnewline
 & Avg. &
0.4445  &
0.3516  &
0.3805  &
0.3001  &
0.3565  &
0.2626  &
0.1682  &
0.1520  &
0.1683  &
0.1506 \tabularnewline
\hline 
\hline 
\multirow{7}{*}{\begin{turn}{90}
Jasper Ridge \#2
\end{turn}} &
\multirow{2}{*}{\emph{Endmembers}} &
\multicolumn{5}{c|}{\textbf{S}pectral \textbf{A}ngle \textbf{D}istance\textbf{ }(\textbf{SAD})} &
\multicolumn{5}{c|}{\textbf{R}oot \textbf{M}ean \textbf{S}quare \textbf{E}rror (\textbf{RMSE})}\tabularnewline
 &  & VCA &
NMF &
$\ell_{1}$-NMF &
$\ell_{1/2}$-NMF &
G-NMF &
VCA &
NMF &
$\ell_{1}$-NMF &
$\ell_{1/2}$-NMF &
G-NMF\tabularnewline
\cline{2-12} 
 & \#1 Tree &
0.2565  &
0.2130  &
0.0680  &
0.0409  &
0.2781  &
0.3268  &
0.1402  &
0.0636  &
0.0707  &
0.1740 \tabularnewline
 & \#2 Water &
0.2474  &
0.2001  &
0.3815  &
0.1682  &
0.2530  &
0.3151  &
0.1106  &
0.0660  &
0.1031  &
0.1375 \tabularnewline
 & \#3 Soil &
0.3584  &
0.1569  &
0.0898  &
0.0506  &
0.3246  &
0.2936  &
0.2557  &
0.2463  &
0.2679  &
0.2840 \tabularnewline
 & \#4 Road &
0.5489  &
0.3522  &
0.4118  &
0.3670  &
0.2776  &
0.2829  &
0.2450  &
0.2344  &
0.2737  &
0.2279 \tabularnewline
 & Avg. &
0.3528  &
0.2305  &
0.2378  &
0.1567  &
0.2833  &
0.3046  &
0.1879  &
0.1526  &
0.1789  &
0.2058 \tabularnewline
\hline 
\hline 
\multirow{7}{*}{\begin{turn}{90}
Jasper Ridge \#3
\end{turn}} &
\multirow{2}{*}{\emph{Endmembers}} &
\multicolumn{5}{c|}{\textbf{S}pectral \textbf{A}ngle \textbf{D}istance\textbf{ }(\textbf{SAD})} &
\multicolumn{5}{c|}{\textbf{R}oot \textbf{M}ean \textbf{S}quare \textbf{E}rror (\textbf{RMSE})}\tabularnewline
 &  & VCA &
NMF &
$\ell_{1}$-NMF &
$\ell_{1/2}$-NMF &
G-NMF &
VCA &
NMF &
$\ell_{1}$-NMF &
$\ell_{1/2}$-NMF &
G-NMF\tabularnewline
\cline{2-12} 
 & \#1 Tree &
0.1168  &
0.0992  &
0.0969  &
0.1081  &
0.1117  &
0.1522  &
0.1000  &
0.1182  &
0.1017  &
0.1164 \tabularnewline
 & \#2 Water &
0.2763  &
0.3124  &
0.0513  &
0.0474  &
0.3369  &
0.1467  &
0.1248  &
0.0418  &
0.0651  &
0.1298 \tabularnewline
 & \#3 Soil &
0.5364  &
0.3748  &
0.3525  &
0.2419  &
0.3414  &
0.2475  &
0.2663  &
0.2536  &
0.2289  &
0.2247 \tabularnewline
 & \#4 Road &
0.4235  &
0.2997  &
0.3906  &
0.4098  &
0.2200  &
0.2105  &
0.2163  &
0.1707  &
0.1834  &
0.1688 \tabularnewline
 & Avg. &
0.3383  &
0.2715  &
0.2228  &
0.2018  &
0.2525  &
0.1892  &
0.1769  &
0.1461  &
0.1448  &
0.1599 \tabularnewline
\hline 
\end{tabular}} 
\par\end{centering}

\centering{}\vspace{0.25cm}
\end{table*}

\begin{table*}[t]
\centering{}\caption{The performances of five benchmark HU algorithms on the Moffett Field\#1
and Moffett Field\#2 hyperspectral images.\label{tab:rst6_MoffettField_1=0000262}}
\vspace{-0.25cm}
 \resizebox{1.75\columnwidth}{!}{%
\begin{tabular}{|l||l|ccccc|ccccc|}
\hline 
\multirow{6}{*}{\begin{turn}{90}
Moffett Field\#1
\end{turn}} &
\multirow{2}{*}{\emph{Endmembers}} &
\multicolumn{5}{c|}{\textbf{S}pectral \textbf{A}ngle \textbf{D}istance\textbf{ }(\textbf{SAD})} &
\multicolumn{5}{c|}{\textbf{R}oot \textbf{M}ean \textbf{S}quare \textbf{E}rror (\textbf{RMSE})}\tabularnewline
 &  & VCA &
NMF &
$\ell_{1}$-NMF &
$\ell_{1/2}$-NMF &
G-NMF &
VCA &
NMF &
$\ell_{1}$-NMF &
$\ell_{1/2}$-NMF &
G-NMF\tabularnewline
\cline{2-12} 
 & \#1 Soil &
0.0760  &
0.0475  &
0.0884  &
0.0717  &
0.1006  &
0.2492  &
0.0923  &
0.0912  &
0.0943  &
0.1085 \tabularnewline
 & \#2 Tree &
0.0418  &
0.0432  &
0.0332  &
0.0357  &
0.0477  &
0.0992  &
0.0540  &
0.0595  &
0.0590  &
0.0842 \tabularnewline
 & \#3 Water &
0.3946  &
0.3528  &
0.3471  &
0.3432  &
0.3697  &
0.2751  &
0.1006  &
0.0790  &
0.0839  &
0.1219 \tabularnewline
 & Avg. &
0.1708  &
0.1478  &
0.1563  &
0.1502  &
0.1727  &
0.2078  &
0.0823  &
0.0766  &
0.0791  &
0.1049 \tabularnewline
\hline 
\hline 
\multirow{6}{*}{\begin{turn}{90}
Moffett Field\#2
\end{turn}} &
\multirow{2}{*}{\emph{Endmembers}} &
\multicolumn{5}{c|}{\textbf{S}pectral \textbf{A}ngle \textbf{D}istance\textbf{ }(\textbf{SAD})} &
\multicolumn{5}{c|}{\textbf{R}oot \textbf{M}ean \textbf{S}quare \textbf{E}rror (\textbf{RMSE})}\tabularnewline
 &  & VCA &
NMF &
$\ell_{1}$-NMF &
$\ell_{1/2}$-NMF &
G-NMF &
VCA &
NMF &
$\ell_{1}$-NMF &
$\ell_{1/2}$-NMF &
G-NMF\tabularnewline
\cline{2-12} 
 & \#1 Soil &
0.0509  &
0.0398  &
0.0360  &
0.0258  &
0.0402  &
0.1120  &
0.1685  &
0.0456  &
0.0703  &
0.1701 \tabularnewline
 & \#2 Tree &
0.0406  &
0.0419  &
0.0684  &
0.0592  &
0.0422  &
0.0403  &
0.0600  &
0.0286  &
0.0417  &
0.0595 \tabularnewline
 & \#3 Water &
0.2793  &
0.3224  &
0.0555  &
0.0726  &
0.3247  &
0.1072  &
0.1771  &
0.0348  &
0.0501  &
0.1779 \tabularnewline
 & Avg. &
0.1236  &
0.1347  &
0.0533  &
0.0525  &
0.1357  &
0.0865  &
0.1352  &
0.0364  &
0.0540  &
0.1359 \tabularnewline
\hline 
\end{tabular}}
\end{table*}

\begin{table*}[t]
\centering{}\caption{The HU results of five benchmark algorithms on the WDM\#1 and WDM\#2
subimages. WDM denotes Washington DC Mall. \label{tab:rst7_WDC_1=0000262=0000263}}
\vspace{-0.25cm}
\resizebox{1.75\columnwidth}{!}{%
\begin{tabular}{|l||l|ccccc|ccccc|}
\hline 
\multirow{9}{*}{\begin{turn}{90}
WDM \#1
\end{turn}} &
\multirow{2}{*}{\emph{Endmembers}} &
\multicolumn{5}{c|}{\textbf{S}pectral \textbf{A}ngle \textbf{D}istance\textbf{ }(\textbf{SAD})} &
\multicolumn{5}{c|}{\textbf{R}oot \textbf{M}ean \textbf{S}quare \textbf{E}rror (\textbf{RMSE})}\tabularnewline
 &  & VCA &
NMF &
$\ell_{1}$-NMF &
$\ell_{1/2}$-NMF &
G-NMF &
VCA &
NMF &
$\ell_{1}$-NMF &
$\ell_{1/2}$-NMF &
G-NMF\tabularnewline
\cline{2-12} 
 & \#1 Roof &
0.18175  &
0.23343  &
0.24818  &
0.22110  &
0.21093  &
0.12021  &
0.16432  &
0.16577  &
0.16744  &
0.17293 \tabularnewline
 & \#2 Grass &
0.24779  &
0.16813  &
0.17883  &
0.17013  &
0.17285  &
0.18837  &
0.19286  &
0.19132  &
0.19167  &
0.19202 \tabularnewline
 & \#3 Water &
0.18327  &
0.17505  &
0.16717  &
0.17077  &
0.17099  &
0.33159  &
0.20351  &
0.20051  &
0.20236  &
0.19769 \tabularnewline
 & \#4 Tree &
0.13241  &
0.15310  &
0.15853  &
0.14407  &
0.15898  &
0.15137  &
0.14405  &
0.14417  &
0.14449  &
0.14406 \tabularnewline
 & \#5 Trail &
0.17445  &
0.26442  &
0.26116  &
0.26343  &
0.27032  &
0.30302  &
0.26970  &
0.27523  &
0.27606  &
0.28636 \tabularnewline
 & \#6 Road &
0.42313  &
0.35212  &
0.34244  &
0.37354  &
0.36481  &
0.36034  &
0.29809  &
0.29570  &
0.30967  &
0.29885 \tabularnewline
 & Avg. &
0.22380  &
0.22437  &
0.22605  &
0.22384  &
0.22481  &
0.24248  &
0.21209  &
0.21212  &
0.21528  &
0.21532 \tabularnewline
\hline 
\hline 
\multirow{10}{*}{\begin{turn}{90}
WDM \#2
\end{turn}} &
\multirow{2}{*}{\emph{Endmembers}} &
\multicolumn{5}{c|}{\textbf{S}pectral \textbf{A}ngle \textbf{D}istance\textbf{ }(\textbf{SAD})} &
\multicolumn{5}{c|}{\textbf{R}oot \textbf{M}ean \textbf{S}quare \textbf{E}rror (\textbf{RMSE})}\tabularnewline
 &  & VCA &
NMF &
$\ell_{1}$-NMF &
$\ell_{1/2}$-NMF &
G-NMF &
VCA &
NMF &
$\ell_{1}$-NMF &
$\ell_{1/2}$-NMF &
G-NMF\tabularnewline
\cline{2-12} 
 & \#1 Asphalt &
0.72505  &
0.81097  &
0.70794  &
0.75846  &
0.83643  &
0.33940  &
0.33944  &
0.33806  &
0.34125  &
0.34041 \tabularnewline
 & \#2 Grass &
1.17554  &
1.41735  &
0.07486  &
1.17832  &
1.44942  &
0.45647  &
0.46033  &
0.33928  &
0.44229  &
0.46026 \tabularnewline
 & \#3 Tree &
0.98158  &
0.69179  &
0.11641  &
0.14168  &
0.63707  &
0.21752  &
0.18320  &
0.23490  &
0.21929  &
0.18256 \tabularnewline
 & \#4 Roof1 &
0.15295  &
0.04880  &
0.03986  &
0.03278  &
0.04388  &
0.24413  &
0.25255  &
0.25370  &
0.26166  &
0.25228 \tabularnewline
 & \#5 Roof2/ &
\multirow{2}{*}{0.18843 } &
\multirow{2}{*}{0.09550 } &
\multirow{2}{*}{1.18046 } &
\multirow{2}{*}{0.31077 } &
\multirow{2}{*}{0.08978 } &
\multirow{2}{*}{0.35982 } &
\multirow{2}{*}{0.38600 } &
\multirow{2}{*}{0.26782 } &
\multirow{2}{*}{0.35865 } &
\multirow{2}{*}{0.38359 }\tabularnewline
 & \ \ \ Shadow &  &  &  &  &  &  &  &  &  & \tabularnewline
 & \#6 Soil &
0.22996  &
0.73903  &
0.80175  &
0.93171  &
0.73604  &
0.26711  &
0.18963  &
0.16798  &
0.16017  &
0.18300 \tabularnewline
 & Avg. &
0.57558  &
0.63391  &
0.48688  &
0.55895  &
0.63210  &
0.31408  &
0.30186  &
0.26696  &
0.29722  &
0.30035 \tabularnewline
\hline 
\end{tabular}}\vspace{-0.25cm}
\end{table*}

\begin{table*}[t]
\begin{centering}
\caption{The performances of five benchmark HU algorithms on the San Diego
Aiport image, which has two versions of ground truths. \label{tab:rst8_SanDiegoAirport_end4=0000265}}
\vspace{-0.25cm}
\resizebox{1.75\columnwidth}{!}{%
\begin{tabular}{|l||l|ccccc|ccccc|}
\hline 
\multirow{8}{*}{\begin{turn}{90}
V1: 4 \emph{endmembers}
\end{turn}} &
\multirow{2}{*}{\emph{Endmembers}} &
\multicolumn{5}{c|}{\textbf{S}pectral \textbf{A}ngle \textbf{D}istance\textbf{ }(\textbf{SAD})} &
\multicolumn{5}{c|}{\textbf{R}oot \textbf{M}ean \textbf{S}quare \textbf{E}rror (\textbf{RMSE})}\tabularnewline
 &  & VCA &
NMF &
$\ell_{1}$-NMF &
$\ell_{1/2}$-NMF &
G-NMF &
VCA &
NMF &
$\ell_{1}$-NMF &
$\ell_{1/2}$-NMF &
G-NMF\tabularnewline
\cline{2-12} 
 & \#1 Grass &
1.0476  &
0.7738  &
0.1811  &
0.0474  &
0.7763  &
0.3164  &
0.3403  &
0.2675  &
0.2692  &
0.3255 \tabularnewline
 & \#2 Tree &
0.1891  &
0.1904  &
0.7613  &
0.8001  &
0.1709  &
0.2257  &
0.2338  &
0.2672  &
0.2851  &
0.2336 \tabularnewline
 & \#3 Roof &
0.3844  &
0.5240  &
0.4924  &
0.4608  &
0.5224  &
0.4508  &
0.3361  &
0.3101  &
0.3491  &
0.3348 \tabularnewline
 & \#4 Ground &
\multirow{2}{*}{0.2187 } &
\multirow{2}{*}{0.2543 } &
\multirow{2}{*}{0.1313 } &
\multirow{2}{*}{0.0609 } &
\multirow{2}{*}{0.2487 } &
\multirow{2}{*}{0.3719 } &
\multirow{2}{*}{0.2593 } &
\multirow{2}{*}{0.2283 } &
\multirow{2}{*}{0.2964 } &
\multirow{2}{*}{0.2562 }\tabularnewline
 & \ \ \ \ /Road &  &  &  &  &  &  &  &  &  & \tabularnewline
 & Avg. &
0.4600  &
0.4356  &
0.3915  &
0.3423  &
0.4296  &
0.3412  &
0.2924  &
0.2683  &
0.2999  &
0.2875 \tabularnewline
\hline 
\multirow{9}{*}{\begin{turn}{90}
V2: 5 \emph{endmembers}
\end{turn}} &
\multirow{2}{*}{\emph{Endmembers}} &
\multicolumn{5}{c|}{\textbf{S}pectral \textbf{A}ngle \textbf{D}istance\textbf{ }(\textbf{SAD})} &
\multicolumn{5}{c|}{\textbf{R}oot \textbf{M}ean \textbf{S}quare \textbf{E}rror (\textbf{RMSE})}\tabularnewline
 &  & VCA &
NMF &
$\ell_{1}$-NMF &
$\ell_{1/2}$-NMF &
G-NMF &
VCA &
NMF &
$\ell_{1}$-NMF &
$\ell_{1/2}$-NMF &
G-NMF\tabularnewline
\cline{2-12} 
 & \#1 Grass &
0.4806  &
0.3382  &
0.3560  &
0.2957  &
0.2544  &
0.3249  &
0.2916  &
0.2964  &
0.2754  &
0.2906 \tabularnewline
 & \#2 Tree &
0.1592  &
0.1361  &
0.1439  &
0.1624  &
0.1430  &
0.2209  &
0.2267  &
0.2212  &
0.2176  &
0.2207 \tabularnewline
 & \#3 Roof &
0.6239  &
0.6509  &
0.6293  &
0.6205  &
0.6409  &
0.3299  &
0.2538  &
0.2626  &
0.2715  &
0.2563 \tabularnewline
 & \#4 Ground &
\multirow{2}{*}{0.2509 } &
\multirow{2}{*}{0.2276 } &
\multirow{2}{*}{0.2200 } &
\multirow{2}{*}{0.2047 } &
\multirow{2}{*}{0.2501 } &
\multirow{2}{*}{0.3443 } &
\multirow{2}{*}{0.2544 } &
\multirow{2}{*}{0.2622 } &
\multirow{2}{*}{0.2566 } &
\multirow{2}{*}{0.2581 }\tabularnewline
 & \ \ \ \ /Road &  &  &  &  &  &  &  &  &  & \tabularnewline
 & \#5 Other &
0.4856  &
0.5421  &
0.5586  &
0.5861  &
0.6157  &
0.3382  &
0.2665  &
0.2735  &
0.2781  &
0.2696 \tabularnewline
 & Avg. &
0.4000  &
0.3790  &
0.3815  &
0.3739  &
0.3808  &
0.3117  &
0.2586  &
0.2632  &
0.2599  &
0.2591 \tabularnewline
\hline 
\end{tabular}}
\par\end{centering}

\centering{}\vspace{0.25cm}
\end{table*}

\section{Experiments Results of 5 Benchmark HU methods on the 18 Hyperspectral
Images\label{sec:ExpRsts}}

In this section, we compare 5 benchmarks HU methods on the 15 hyperspectral
images summarized in Section\ \ref{sec:Real-Hyperspectral-Images}.
The selected 5 methods are widely compared against the state-of-the-art
HU methods. The aim is to provide benchmark HU results, saving other's
effort to evaluate their new HU methods.

\subsection{Five Benchmark Methods Compared in this Section\label{sub:Five-Benchmark-Methods}}

\textbf{1)}. Vertex Component Analysis~\cite{Jose_05_TGRS_Vca} (VCA)
is a classic geometric method. The code is available on \url{http://www.lx.it.pt/bioucas/code.htm}. 

\textbf{2)}. Nonnegative Matrix Factorization~\cite{Lee_99_Nature_NMF}
(NMF) is a benchmark statistical method. The code is obtained from
\url{http://www.cs.helsinki.fi/u/phoyer/software.html}. 

\textbf{3)}. Nonnegative sparse coding~\cite{Hoyer_02_NNSP_NMF_l1}
($\ell_{1}$-NMF) is a classic sparse regularized NMF method. The
code is download from \url{http://www.cs.helsinki.fi/u/phoyer/software.html}. 

\textbf{4)}. $\ell_{1/2}$ \emph{sparsity-constrained} NMF~\cite{Qian_11_TGRS_NMF+l1/2}
($\ell{}_{\text{1/2}}$-NMF) is a state-of-the-art method that could
get sparser results than $\ell_{1}$-NMF. Since the code is unavailable,
we implement it.

\textbf{5)}. Graph regularized NMF~\cite{Cai_11_PAMI_GNMF,JmLiu_12_SlectedTopics_W-NMF}
(G-NMF) is an interesting algorithm that transfers graph information
inherent in the hyperspectral image into the \emph{abundance} space.

We will share the code for all the above five algorithms.

\subsection{Experiment Settings and Experiment Results}

Generally, there are one or two hyper-parameters in the 5 benchmark
methods. We use the coarse to fine grid search method to find the
best hyper-parameters\ \cite{fyzhu_2014_IJPRS_SSNMF,fyzhu_2014_JSTSP_RRLbS,fyzhu_2014_TIP_DgS_NMF}.
The initialization of the \emph{endmembers} $\mtbfM$ and \emph{abundances}
$\mtbfA$ is very important to the final HU results\ \cite{nWang_13_SelectedTopics_EDC-NMF}.
 We employ the same benchmark algorithm, i.e., VCA\ \cite{Jose_05_TGRS_Vca},
to initialize $\mtbfM$ and $\mtbfA$ for the five methods on all
the 15 datasets. There are two advantages: (1) it is helpful to generate
the reproducible results by the VCA; (2) VCA is really simple, effective
and solid. 

Two evaluation metrics are used to obtain the quantitative comparison
results: (1) the Spectral Angle Distance (SAD)\ \cite{xiaoqiangLu_2013_TGRS_ManifoldSparseNMF,LiuXueSong_2011_TGRS_ConstrainedNMF,fyzhu_2014_IJPRS_SSNMF,fyzhu_2014_JSTSP_RRLbS,fyzhu_2014_TIP_DgS_NMF}
is used to verify the performance of the estimated \emph{endmembers};
(2) the Root Mean Square Error (RMSE)\ \cite{Qian_11_TGRS_NMF+l1/2,fyzhu_2014_IJPRS_SSNMF,Kelly_2011_TGRS_SpatiAdaptiveUnmixing,fyzhu_2014_JSTSP_RRLbS,fyzhu_2014_TIP_DgS_NMF}
is for the evaluation of the estimated \emph{abundances}. To get a
valid results, each experiment is repeated 50 times and the mean result
is provided in the Tables\ \ref{tab:rst1_Samson_1=0000262=0000263},\ \ref{tab:rst2_Cuprite},\ \ref{tab:rst3_Urban_end4,5,6},\ \ref{tab:rst4_Urban_R1=000026R2},\ \ref{tab:rst5_JasperRidge_1=0000262=0000263},\ \ref{tab:rst6_MoffettField_1=0000262},\ \ref{tab:rst7_WDC_1=0000262=0000263},\ \ref{tab:rst8_SanDiegoAirport_end4=0000265}.
In short, we summarize the tables that display the results on all
the 15 datasets:

\textbf{1)}. Table\ \ref{tab:rst1_Samson_1=0000262=0000263} summarizes
the experiment results of all the five benchmark algorithms (reviewed
in Section\ \ref{sub:Five-Benchmark-Methods}). There are three tables,
displaying the HU results on the three subimages, i.e., Samson\#1,
Samson\#2 and Samson\#3 respectively.

\textbf{2)}. In Table\ \ref{tab:rst2_Cuprite}, the HU results on
the Cuprite image are summarized. Note that the ground truth \emph{abundance}
is not available; only the \emph{endmember} results are compared. 

\textbf{3)}. Table\ \ref{tab:rst3_Urban_end4,5,6} illustrates the
HU results of 5 benchmark algorithms on the full HYDICE Urban image.
There are 3 versions of ground truths for the Urban images. Accordingly,
we summarize 3 versions of HU results in the 3 sub-tables.

\textbf{4)}. In Table\ \ref{tab:rst4_Urban_R1=000026R2}, the HU
results on the two subimages, Urban\#1 and Urban\#2, are summarized.
There are six \emph{endmembers} as shown in the Table\ \ref{tab:rst4_Urban_R1=000026R2}.

\textbf{5)}. In Table\ \ref{tab:rst5_JasperRidge_1=0000262=0000263},
the experiment results on the 3 subimages from the JasperRidge is
displayed. The 3 subimage are JasperRidge\#1, JasperRidge\#2 and JasperRidge\#3. 

\textbf{6)}. Table\ \ref{tab:rst6_MoffettField_1=0000262} illustrates
the HU results on the 2 subimags selected from Moffett Field. They
are Moffett Field\#1 and Moffett Field\#2 respectively as shown in
the Table.

\textbf{7)}. Table\ \ref{tab:rst7_WDC_1=0000262=0000263} illustrates
the HU results on the 2 subimags selected from the Washington DC Mall.
They are WDM\#1 and WDM\#2 respectively.

\textbf{8)}. Table\ \ref{tab:rst8_SanDiegoAirport_end4=0000265}
displays the HU results on the two versions of ground truths on the
San Diego Airport dataset.

\section{Conclusion}

Hyperspectral unmixing (HU) is an important preprocessing step for
a great number of hyperpsectral applications. To promote the HU research,
we accomplished this paper by emphasizing on the generation of standard
HU datasets (with ground truths) and the summarization of benchmark
results of the state-of-the-art HU algorithms. Specifically, this
is the first paper to propose a general labeling method for the HU.
Via it, we summarized and labeled up to 15 hyperpsectral images. To
further enrich the HU datasets, an interesting method has been proposed
to transform the hyperspectral classification datasets for the HU
research. Besides, we reviewed and implemented a widely accepted algorithm
to generate a very complex simulated hyperpsectral image for the HU
study. Such synthetic dataset is helpful to verify the HU algorithms
under four different conditions. 

To summarize the benchmark HU results, we reviewed up to 10 state-of-the-art
HU algorithms, and selected the five most benchmark HU algorithms
for the comparison. Those HU algorithms are compared on the 15 hyperpsectral
images. To the best of our knowledge, this is the first paper to compare
the benchmark algorithms on so many real hyperpsectral images. The
experient results on this paper may provide a valid baseline for the
evaluation of new HU algorithms. 

{\scriptsize{}\bibliographystyle{IEEEtran}
\bibliography{18F__Backups_Dropbox_20150901_Dropbox_Dropbox_s___rUnmixingDatasets_ClassicAlgs_referenceBib2}
}{\scriptsize \par}
\end{document}